\documentclass{fundam}

\usepackage{url} 
\usepackage{graphicx}
\usepackage{tikz}
\usepackage{amsfonts}
\usetikzlibrary{automata, positioning, arrows}
\usepackage{mathptmx}
\usepackage[lined,linesnumbered,boxed]{algorithm2e}
\usepackage{caption}
\usepackage{subcaption}
\usepackage{float}
\usepackage{multirow}

\begin{document}

%
\setcounter{page}{1}
\publyear{2021}
\papernumber{0001}
\volume{178}
\issue{1}
%

\title{The Archerfish Hunting Optimizer: a novel metaheuristic algorithm for global optimization}

\address{harous{@}uaeu.ac.ae}

\author{Farouq Zitouni\\
Department of Computer Science, Kasdi Merbah University -- Ouargla\\
LIRE Laboratory, Abdelhamid Mehri University -- Constantine\\
Algeria\\
farouq.zitouni{@}univ-constantine2.dz
\and 
Saad Harous\\
Department of Computer Science and Software Engineering UAE University\\
Abu Dhabi, United Arab Emirates\\
harous{@}uaeu.ac.ae
\and Abdelghani Belkeram\\
Department of Computer Science, Kasdi Merbah University\\
Ouargla, Algeria\\
abd001979{@}gmail.com
\and Lokman Elhakim Baba Hammou\\
Department of Computer Science, Kasdi Merbah University\\
Ouargla, Algeria\\
lokman.babahamou6{@}gmail.com
}

\maketitle

\runninghead{F. Zitouni, S. Harous, A. Belkeram, and LE. Baba Hammou}{AHO}

\begin{abstract}
  Global optimization solves real-world problems numerically or analytically by minimizing their objective functions. Most of the analytical algorithms are greedy and computationally intractable. Metaheuristics are nature-inspired optimization algorithms. They numerically find a near-optimal solution for optimization problems in a reasonable amount of time. We propose a novel metaheuristic algorithm for global optimization. It is based on the shooting and jumping behaviors of the archerfish for hunting aerial insects. We name it the Archerfish Hunting Optimizer (AHO). We Perform two sorts of comparisons to validate the proposed algorithm's performance. First, AHO is compared to the 12 recent metaheuristic algorithms (the accepted algorithms for the 2020's competition on single objective bound-constrained numerical optimization) on ten test functions of the benchmark CEC 2020 for unconstrained optimization. Second, the performance of AHO and 3 recent metaheuristic algorithms, is evaluated using five engineering design problems taken from the benchmark CEC 2020 for non-convex constrained optimization. The experimental results are evaluated using the Wilcoxon signed-rank and the Friedman tests. The statistical indicators illustrate that the Archerfish Hunting Optimizer has an excellent ability to accomplish higher performance in competition with the well-established optimizers.
\end{abstract}


\section{Introduction}

Approximation algorithms were proposed for the first time in the 1960s to solve challenging optimization problems \cite{vazirani2013approximation}. They are called approximation algorithms because they generate near-optimal solutions. They were mainly used to solve optimization problems that could not be solved efficiently using computational techniques available at that period \cite{grossman1997computational}. The $NP$-completeness theory also had a considerable contribution to approximation algorithms' maturing since the need to solve $NP$-hard optimization problems became the most priority to deal with the computational intractability \cite{hochba1997approximation}. Some optimization problems are easy to solve (i.e., generating near-optimal solutions is quick), while for other ones, that task is as hard as finding optimal solutions \cite{agarwal2002exact}.

Approximation algorithms using probabilistic and randomized techniques had known tremendous advances between the 1980s and 1990s. They were named metaheuristic algorithms \cite{glover1986future}. Famous metaheuristic algorithms might include, for example, Simulated Annealing \cite{van1987simulated}, Ant Colony Optimization \cite{dorigo1999ant}, Evolutionary Computation \cite{fogel1998evolutionary}, Tabu Search \cite{glover1990tabu}, Memetic Algorithms \cite{moscato1989evolution}, and Particle Swarm Optimization \cite{kennedy1995particle}, to name but a few. During the past three decades, many metaheuristic algorithms have been proposed in the literature. Most of them have been assessed experimentally and have shown good performance for solving real-world optimization problems \cite{hussain2017common}. Metaheuristic algorithms aim to find the best possible solutions and guarantee that such solutions satisfy some criteria \cite{ezugwu2019mathematical}. The No-Free-Lunch theorem \cite{ho2002simple} proves that universal metaheuristic algorithms for solving all the optimization problems are non-existent, which justifies the growing amount of the proposed state-of-the-art. In other words, if a certain metaheuristic algorithm efficiently solves some optimization problems, it will systematically show mediocre performance for other ones. Also, the theorem states that all metaheuristic algorithms' averaged performance on all the optimization problems is the same. Tables \ref{table1} and \ref{table2} encompass some well-known metaheuristic algorithms (these algorithms are split into four families, such as evolutionary-based, swarm-based, physical-based, and human-based algorithms \cite{9310185}).

\begin{table}[!]
\centering
\caption{Popular evolutionary-based and swarm-based metaheuristic algorithms.}
\label{table1}
\resizebox{\textwidth}{!}{
\begin{tabular}{ll}
\hline
\multicolumn{1}{c}{\textbf{Evolutionary-based metaheuristic algorithms}} & \multicolumn{1}{c}{\textbf{Swarm-based metaheuristic algorithms}} \\ \hline
Genetic algorithm \cite{holland1992genetic} & Particle swarm optimization \cite{kennedy2010encyclopedia} \\
Evolution strategies \cite{bergmann2012optimization} & Ant colony optimization \cite{dorigo1999ant} \\
Evolutionary programming \cite{fogel1966artificial} & Artificial bee colony \cite{karaboga2007powerful} \\
Genetic programming \cite{koza1992genetic} & Grey wolf optimizer \cite{mirjalili2014grey} \\
Differential evolution \cite{storn1997differential} & Bat algorithm \cite{yang2010new} \\
Biogeography-based optimization \cite{simon2008biogeography} & Whale optimization algorithm \cite{mirjalili2016whale} \\
Covariance matrix adaptation evolution strategy \cite{hansen2006cma} & Dragonfly algorithm \cite{mirjalili2016dragonfly} \\
Quantum-inspired evolutionary algorithm \cite{talbi2017new} & Dolphin echolocation \cite{kaveh2013new} \\
 & Fruit fly optimization \cite{pan2012new} \\
 & Krill herd \cite{gandomi2012krill} \\
 & Bird mating optimizer \cite{askarzadeh2013new} \\
 & Hunting search \cite{oftadeh2010novel} \\
 & Firefly algorithm \cite{yang2010firefly} \\
 & Dolphin partner optimization \cite{shiqin2009dolphin} \\
 & Cuckoo search \cite{yang2009cuckoo} \\
 & Social spider optimization \cite{james2015social} \\
 & Bee collecting pollen algorithm \cite{lu2008novel} \\
 & Marriage in honey bees \cite{abbass2001mbo} \\
 & Monkey search \cite{mucherino2007monkey} \\
 & Termite \cite{roth2005termite} \\
 & Fish swarm algorithm \cite{li2003new} \\
 & Grasshopper optimisation algorithm \cite{saremi2017grasshopper} \\
 & Seagull optimization algorithm \cite{dhiman2019seagull} \\
 & Salp swarm algorithm \cite{mirjalili2017salp} \\
 & Selfish herd optimizer \cite{cuevas2020selfish} \\
 & Moth-flame optimization algorithm \cite{mirjalili2015moth} \\
 & Ant lion optimizer \cite{mirjalili2015ant} \\
 & Harris hawks optimization \cite{heidari2019harris} \\
 & Slime mould algorithm \cite{li2020slime} \\
 & Moth search algorithm \cite{wang2018moth} \\
 & Elephant herding optimization \cite{wang2015elephant} \\
 & Earthworm optimisation algorithm \cite{wang2018earthworm} \\
 & Monarch butterfly optimization algorithm \cite{wang2019monarch} \\
 & Rooted tree optimization algorithm \cite{labbi2016new} \\
 & Tunicate swarm algorithm \cite{kaur2020tunicate} \\ \hline
\end{tabular}
}
\end{table}

\begin{table}[!]
\centering
\caption{Popular physical-based and human-based metaheuristic algorithms.}
\label{table2}
\resizebox{\textwidth}{!}{
\begin{tabular}{ll}
\hline
\multicolumn{1}{c}{\textbf{Physical-based metaheuristic algorithms}} & \multicolumn{1}{c}{\textbf{Human-based metaheuristic algorithms}} \\ \hline
Simulated annealing \cite{kirkpatrick1983optimization} & Teaching learning-based optimization \cite{rao2011teaching} \\
Thermodynamic laws \cite{vcerny1985thermodynamical} & Harmony search \cite{geem2001new} \\
Gravitation \cite{webster2003local,rashedi2009gsa} & Taboo search \cite{fogel1998artificial} \\
Big bang--big crunch\cite{erol2006new} & Group search optimizer \cite{he2006novel} \\
Charged system \cite{kaveh2010novel} & Imperialist competitive algorithm \cite{atashpaz2007imperialist} \\
Central force \cite{formato2007central} & League championship algorithm \cite{kashan2009league} \\
Chemical reaction \cite{alatas2011acroa} & Colliding bodies optimization \cite{kaveh2014colliding} \\
Black hole \cite{hatamlou2013black} & Interior search algorithm \cite{gandomi2014interior} \\
Ray \cite{kaveh2012new} & Mine blast algorithm \cite{sadollah2013mine} \\
Small-world \cite{du2006small} & Soccer league competition algorithm \cite{moosavian2014soccer} \\
Galaxy-based \cite{shah2011principal} & Seeker optimization algorithm \cite{dai2009seeker} \\
General relativity theory \cite{moghaddam2012curved} & Social-based algorithm \cite{ramezani2013social} \\
Sine cosine algorithm \cite{mirjalili2016sca} & Exchange market algorithm \cite{ghorbani2014exchange} \\
Multi-verse optimizer \cite{mirjalili2016multi} & Nomadic people optimizer \cite{salih2020new} \\
Inclined planes system optimization \cite{mozaffari2016ipo} & Group counseling optimization algorithm  \cite{eita2014group} \\
Firework algorithm \cite{tan2010fireworks} &  \\
Modified inclined planes system optimization \cite{mohammadi2017iir} &  \\
Simplified inclined planes system optimization \cite{mohammadi2019simplified} &  \\
Spherical search \cite{kumar2019spherical} &  \\
Solar system algorithm \cite{9310185} &  \\ \hline
\end{tabular}
}
\end{table}

Mainly, there are two groups of metaheuristic algorithms: population-based and individual-based algorithms \cite{boussaid2013survey}. Population-based algorithms use several agents, whereas individual-based algorithms use one agent. In the first group, several individuals swarm in the search space and cooperatively evolve towards the global optimum \cite{beheshti2013review}. In the second group, one individual moves in the search space and evolves towards the global optimum \cite{dougan2015new}. In both groups, individuals' positions are randomly initialized in the search space and are modified over generations until the satisfaction of specific criteria \cite{boussaid2013survey}.

Metaheuristic algorithms have two fundamental components: exploration and exploitation \cite{hussain2019exploration}. The exploration is called global optimization or diversification. The exploitation is named local optimization or intensification. The exploration allows metaheuristic algorithms to discover new search space regions and avoid being trapped in local optimums \cite{xu2014exploration}. The exploitation permits metaheuristic algorithms to zoom on a particular area to find the best solution \cite{xu2014exploration}. Any metaheuristic algorithm should find the best balance between diversification and intensification; otherwise, the found solutions' quality is compromised \cite{yang2014metaheuristic}. Too many exploration operations may result in a considerable waste of effort: i.e., the algorithm jumps from one location to another without concentrating on enhancing the current solution's quality \cite{salleh2018exploration}. Excessive exploitation operations may lead the algorithm to be trapped in local optimums and to converge prematurely \cite{morales2020better}. The main weakness of metaheuristic algorithms is the sensitivity to the tuning of controlling parameters. Also, the convergence to the global optimum  is not always guaranteed \cite{dreo2006metaheuristics}.

We propose a novel swarm-based metaheuristic algorithm for global optimization, named the Archerfish Hunting Optimizer (AHO). The algorithm is inspired by the shooting and jumping behaviors of archerfish when catching prey. The prominent features of AHO are:

\begin{itemize}
  \item AHO has three controlling parameters to set, the population size, the swapping angle between the exploration and exploitation phases, and the attractiveness rate between the archerfish and the prey.
  \item AHO uses elementary laws of physics (i.e., equations of the general ballistic trajectory) to determine the positions of new solutions.
  \item The swapping angle controls the balance between the exploration and exploitation of the search space.
\end{itemize}

The performance of AHO is assessed using the benchmark CEC 2020 for unconstrained optimization. The considered benchmark contains ten challenging single objective test functions. For further information on this benchmark, the reader is referred to \cite{yue2019problem}. The obtained results are compared to 12 most recent state-of-the-art metaheuristic algorithms (the accepted algorithms for the 2020's competition on single objective bound-constrained numerical optimization). In addition, the performance of AHO is evaluated on five engineering design problems selected from the benchmark CEC 2020 for non-convex constrained optimization. More details on this benchmark is available in \cite{kumar2020test}. The collected results are opposed to 3 most recent state-of-the-art metaheuristic algorithms. The experimental outcomes are judged using the Wilcoxon signed-rank and the Friedman tests. The statistical results show that AHO produces very encouraging and most of the times competitive results compared to the well-established metaheuristic methods.

The rest of the paper is organized as follows. Section \ref{section2} illustrates the hunting behavior of archerfish and provides the source of inspiration for AHO. Section \ref{section3} describes the proposed metaheuristic algorithm and its mathematical model. Sections \ref{section4} and \ref{section5} represent and discuss the statistical results and comparative study on some unconstrained and constrained optimization problems. Section \ref{section6} sums up the paper and concludes with some future directions.

\section{Source of inspiration}
\label{section2}

\begin{figure}[!]
\centering
\fbox{
\includegraphics[scale=0.20]{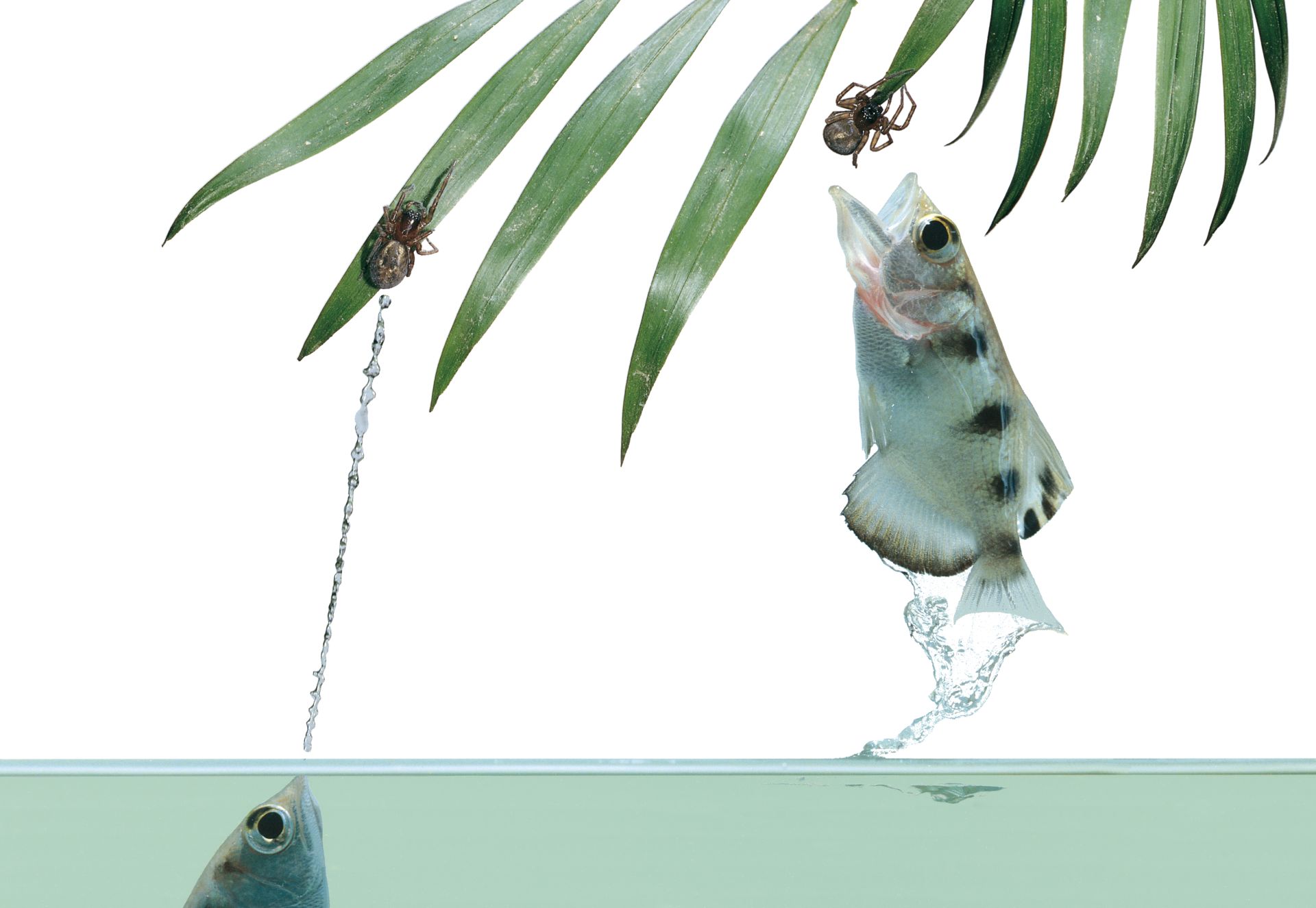}
}
\caption{The hunting mechanisms of archerfish.}
\label{figure1}
\end{figure}

The archerfish form a monotypic family, called Toxotes chatareus. They mainly live in mangrove areas of the Indo Pacific \cite{luling1963archer}. They possess one of the most complex and exciting feeding behaviors: prey aerial insects by shooting them down with water droplets spit from their mouths. Figure \ref{figure1} shows the shape of an archerfish and its hunting mechanisms. An archerfish uses two ways to capture insects: i) it dislodges the target with a powerful jet of water (left archerfish), or ii) it jumps at the prey if the latter is close enough (right archerfish) \cite{rossel2002predicting}.

In practice, the shooting technique is less tiring than jumping. It permits many consecutive shots. However, retrieving a fallen prey is uncertain because other archerfish might steal it \cite{schuster2006animal,schuster2004archer}. When jumping, an archerfish locates itself directly below the target \cite{shih2017archer}. When shooting, an archerfish takes a more lateral position \cite{vailati2012archer}. Archerfish eject water droplets at aerial insects, beating them onto the water surface to be eaten. Since the archerfish's eyes remain entirely below the water surface during the sighting and spitting, it needs to deal with refraction effects at the air-water interface \cite{dill1977refraction}. In this paper, we adopt the following principles to outline the instructions of AHO.

\begin{itemize}
  \item Archerfish live in a flock (AHO is a population-based metaheuristic algorithm).
  \item Archerfish use two different hunting mechanisms: shooting and jumping (exploration and exploitation phases).
  \item Archerfish can steal captured prey from each other (cooperative search and information sharing).
  \item Swapping between jumping and shooting behaviors is controlled by the perceiving angle (balancing between exploration and exploitation phases).
\end{itemize}

\section{Archerfish hunting optimizer}
\label{section3}

We illustrate the exploration and exploitation phases of the proposed AHO, inspired by archerfish' shooting and jumping behaviors when hunting insects. AHO is a gradient-free optimization method that can solve any optimization problem with a proper formulation of the objective function. We assume a search space of dimension $d$ that contains several archerfish. The flock size (i.e., the number of archerfish) is $N$, and the location of archerfish $i$ at iteration $t$ is given as follows.
\[X^{\langle i,t \rangle}=(x_1,x_2,\ldots,x_d)\]

Each entry of $X^{\langle i,t \rangle}$ has a range of allowed values: i.e., $X^{\langle i,t \rangle}=(x_j) \in [x_{j}^{\min},x_{j}^{\max}]$ (where $i \in \{1,\ldots,N\}$ and $j \in \{1,\ldots,d\}$). At iteration $t=0$, the location $X^{\langle i,0 \rangle}$ is initialized randomly using Equation \ref{equation1}.

 \begin{equation}
\label{equation1}
X^{\langle i,0 \rangle}=(\alpha_1 \times (x^{\max}_{1}-x^{\min}_{1})+x^{\min}_{1},\ldots,\alpha_d \times (x^{\max }_{d}-x^{\min}_{d})+x^{\min}_{d})
\end{equation}

\begin{tabular}{lll}
where &  & \\
$\alpha_1,\ldots,\alpha_d$ & : & Uniformly distributed random numbers between 0 and 1. \\
\end{tabular}

\begin{figure}[!]
\centering
\fbox{
\includegraphics[scale=0.30]{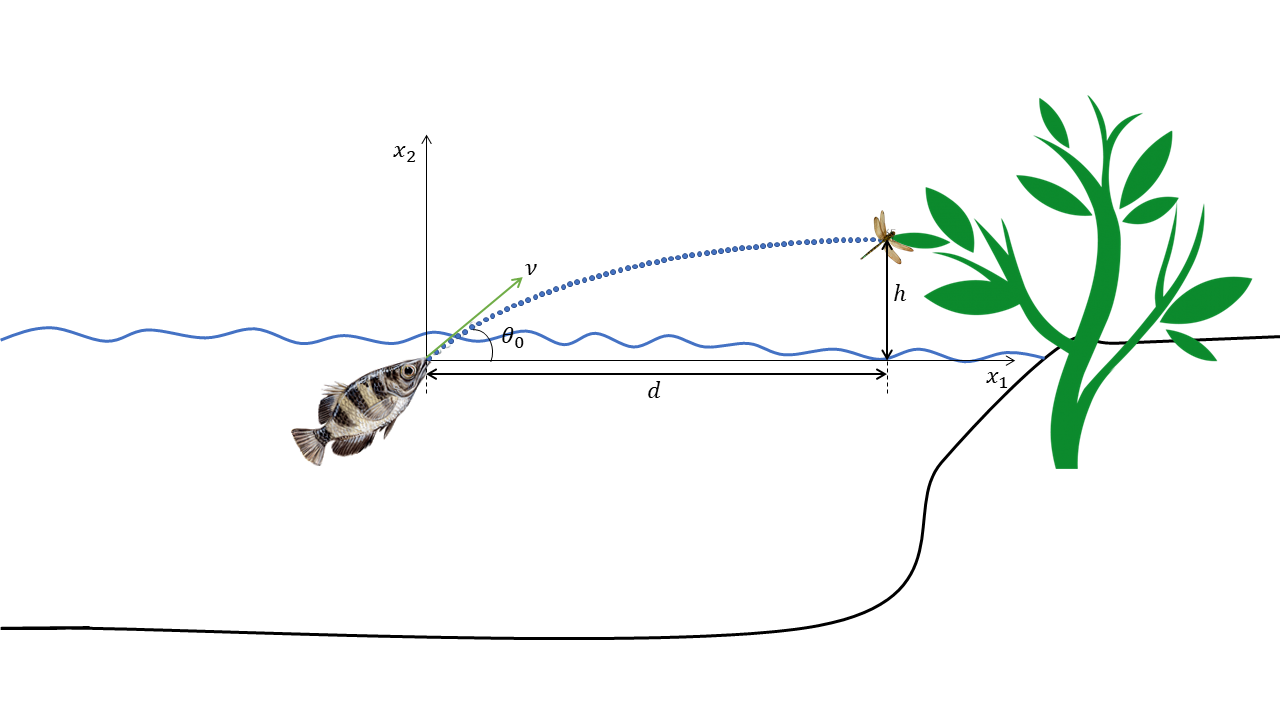}
}
\caption{Shooting behavior of the archerfish.}
\label{figure2}
\end{figure}

Figure \ref{figure2} depicts the shooting behavior of an archerfish (exploration of the search space). The water droplet motion is modeled using equations of the general ballistic trajectory \cite{wheelon1959free}. It is determined by the acceleration of gravity ($g$), the launch speed ($\nu$), and the perceiving angle ($\theta_0$), provided that the air friction is negligible. We suppose that the prey (i.e., dragonfly) is located at the trajectory diagram's peak. When the insect is shot by an archerfish $k$, it will vertically fall onto the water surface. When an archerfish $i$ senses the vibrations initiated by the prey, it moves towards its location using Equation \ref{equation2}.

\begin{equation}
\label{equation2}
X^{\langle i,t+1 \rangle}=X^{\langle i,t \rangle} + e^{-\|X^{\langle k,t \rangle}_{prey}-X^{\langle i,t \rangle}\|^2}(X^{\langle k,t \rangle}_{prey}-X^{\langle i,t \rangle})
\end{equation}

\begin{tabular}{lll}
where &  & \\
$X^{\langle i,t+1 \rangle}$ & : & The next location of archerfish $i$. \\
$X^{\langle i,t \rangle}$ & : & The current location of archerfish $i$. \\
$\|.\|$ & : & The Euclidean distance. \\
$X^{\langle k,t \rangle}_{prey}$ & : & The prey's location. It is computed using Equation \ref{equation3}. \\
$\epsilon$ & : & \begin{tabular}[c]{@{}l@{}}A vector of random numbers generated by a uniform distribution.\\It represents refraction effects at the air-water interface.\end{tabular} \\
$X^{\langle k,t \rangle}$ & : & The location of archerfish $k$, which has shot the insect. \\
\end{tabular}

\begin{equation}
\label{equation3}
X^{\langle k,t \rangle}_{prey}=X^{\langle k,t \rangle}+(0,\ldots,\frac{\nu^2}{2g} \times \sin2\theta_0,\ldots,0)+\epsilon
\end{equation}

The position of the entry given the by term $\frac{\nu^2}{2g} \times \sin2\theta_0$ is a random number in the range $\{1,\ldots,d\}$. For simplicity, fraction $\frac{\nu^2}{2g}$ will be replaced by variable $\omega$. It defines the attractiveness rate of an archerfish to a specific prey.

\begin{figure}[!]
\centering
\fbox{
\includegraphics[scale=0.30]{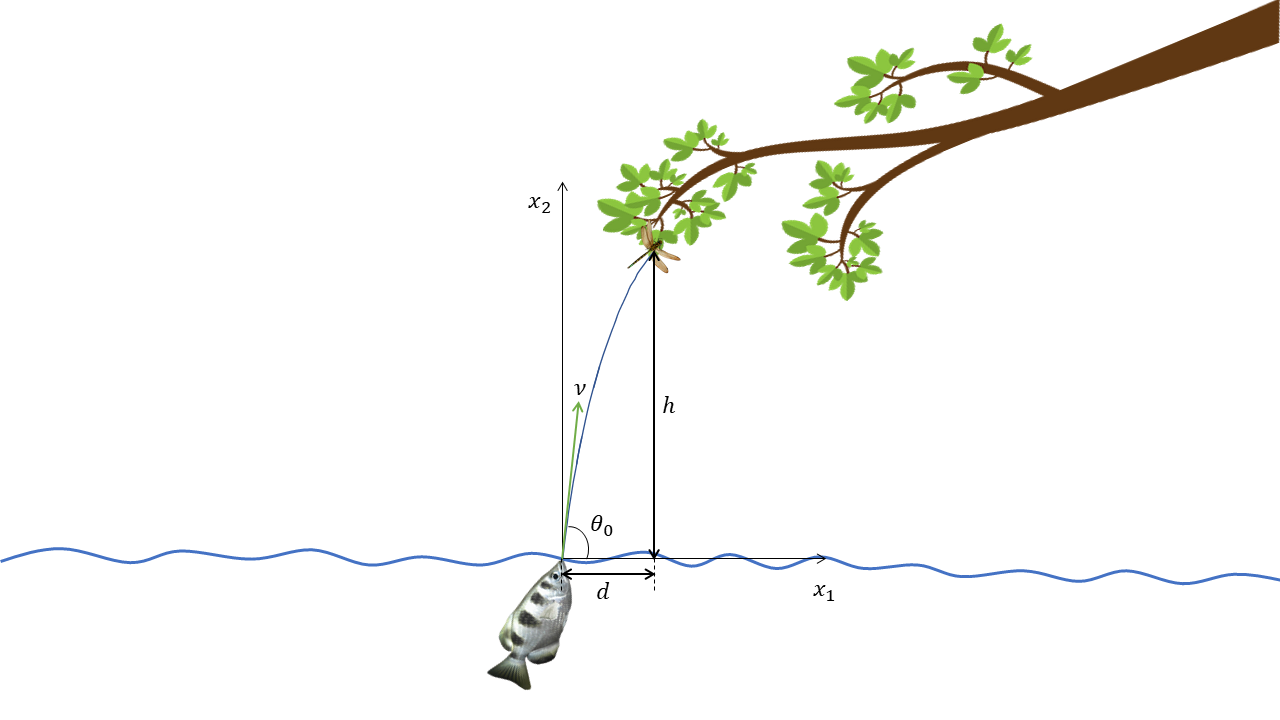}
}
\caption{Jumping behavior of the archerfish.}
\label{figure3}
\end{figure}

Figure \ref{figure3} describes the jumping behavior of an archerfish (exploitation of the search space). The archerfish jumps at the prey and catch it. Similarly, the motion of the archerfish is defined by the acceleration of gravity ($g$), its launch speed ($\nu$), and its perceiving angle ($\theta_0$), provided that the air friction is negligible. We suppose that the prey (i.e., dragonfly) is located at the trajectory diagram's peak. When an archerfish $i$ decides to capture an insect, it moves towards its location using Equation \ref{equation4}.

\begin{equation}
\label{equation4}
X^{\langle i,t+1 \rangle}=X^{\langle i,t \rangle} + e^{-\|X^{\langle i,t \rangle}_{prey}-X^{\langle i,t \rangle}\|^2}(X^{\langle i,t \rangle}_{prey}-X^{\langle i,t \rangle})
\end{equation}

\begin{tabular}{lll}
where &  & \\
$X^{\langle i,t+1 \rangle}$ & : & The next location of archerfish $i$. \\
$X^{\langle i,t \rangle}$ & : & The current location of archerfish $i$. \\
$\|.\|$ & : & The Euclidean distance. \\
$X^{\langle i,t \rangle}_{prey}$ & : & The prey's location. It is computed using Equation \ref{equation5}. \\
$\epsilon$ & : & \begin{tabular}[c]{@{}l@{}}A vector of random numbers drawn from a uniform distribution.\\It represents refraction effects at the air-water interface.\end{tabular} \\
\end{tabular}

\begin{equation}
\label{equation5}
X^{\langle i,t \rangle}_{prey}=X^{\langle i,t \rangle}+(0,\ldots,\frac{\nu^2}{2g} \times \sin2\theta_0,\ldots,\frac{\nu^2}{2g} \times \sin^2\theta_0,\ldots,0)+\epsilon
\end{equation}

The positions of the entries given by the terms $\frac{\nu^2}{2g} \times \sin2\theta_0$ and $\frac{\nu^2}{2g} \times \sin^2\theta_0$ are mandatory distinct random numbers in the range $\{1,\ldots,d\}$. For simplicity, fraction $\frac{\nu^2}{2g}$ will be replaced by variable $\omega$. It defines the attractiveness rate of an archerfish to a specific prey.

\begin{figure}[!]
\centering
\fbox{
\includegraphics[scale=0.25]{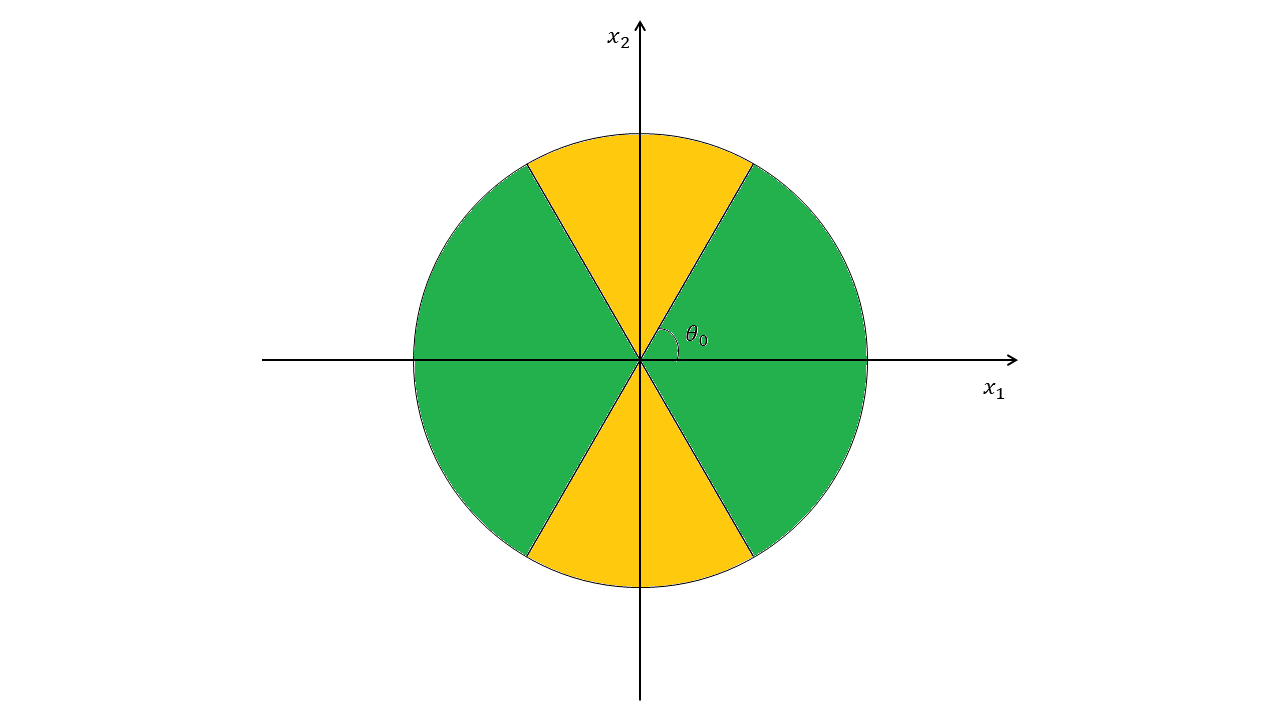}
}
\caption{Swapping between exploration and exploitation.}
\label{figure4}
\end{figure}

The value of perceiving angle ($\theta_0$) guarantees the swapping between the exploration and exploitation phases. Figure \ref{figure4} delimits the ranges of perceiving angles, where AHO is supposed to explore (green areas) or exploit (orange regions) the search space. Hence, the closer the value of $\theta_0$ is to $\frac{\pi}{2}$ or $-\frac{\pi}{2}$, the better AHO tends to exploit the search space and vice versa. The value of $\theta_0$ is generated randomly using Equation \ref{equation6}.

\begin{equation}
\label{equation6}
\theta_0=(-1)^b \times \alpha \times \pi
\end{equation}

\begin{tabular}{lll}
where &  & \\
$b \sim \mathfrak{B}(0.5)$ & : & Bernoulli distribution (success' probability equal to 0.5) \cite{cramer2004random}. \\
$\alpha$ & : & Uniformly distributed random number between 0 and 1. \\
$\pi$ & : & Archimedes' constant equal to 3.14. \\
\end{tabular}

To avoid getting trapped in local optimums, AHO uses a simple strategy. Suppose a given archerfish location $X^{\langle i,t \rangle}$ at iteration $t$ is not enhanced for a fixed number of iterations (e.g., $d \times N$). In this case, the corresponding archerfish moves to a new place according to a Lévy Flight \cite{yang2010firefly}. The new location of $X^{\langle i,t \rangle}$ is generated using Equations \ref{equation7} and \ref{equation8}. Algorithm \ref{algorithm1} shows the pseudo-code of AHO.

\begin{equation}
\label{equation7}
X^{\langle i,t+1 \rangle}=X^{\langle i,t \rangle}+\alpha\bigg[\frac{u_1}{(v_1)^{1/\beta}},\ldots,\frac{u_d}{(v_d)^{1/\beta}}\bigg]
\end{equation}

\begin{equation}
\label{equation8}
\left\{
\begin{array}{lllll}
u_i \sim \mathfrak{N}(0,\sigma^2) & , & \sigma=\bigg(\frac{\Gamma(1+\beta)\sin(\frac{\pi\beta}{2})}{\Gamma(\frac{1+\beta}{2}) \times \beta \times 2^{\frac{\beta-1}{2}}}\bigg)^\frac{1}{\beta} & , & i \in \{1,\ldots,d\} \\
v_i \sim \mathfrak{N}(0,\acute{\sigma}^2) & , & \acute{\sigma}=1 & , & i \in \{1,\ldots,d\} \\
\end{array}
\right.
\end{equation}

\begin{tabular}{lll}
where &  & \\
$\Gamma$ & : & Gamma function \cite{artin2015gamma}. \\
$\mathfrak{N}(\mu,\sigma)$ & : & Normal distribution of mean $\mu$ and standard deviation $\sigma$ \cite{cramer2004random}. \\
$\beta$ & : & The power law index  ($\beta=1.5$). \\
$\alpha$ & : & Uniformly distributed random number between 0 and 1. \\
\end{tabular}

\begin{algorithm}[!]
\SetAlgoLined

\KwIn{$d$ (the dimension of the search space).}
\KwIn{$[x_{1}^{\min},x_{1}^{\max}], \ldots, [x_{d}^{\min},x_{d}^{\max}]$ (the decision variables' domains).}
\KwIn{$f$ (the objective function to be minimized).}
\KwIn{$\theta$ (the swapping angle between the exploration and exploitation phases).}
\KwIn{$\omega$ (the attractiveness rate).}

\BlankLine

\For{$i\gets 1$ \KwTo $N$}{
    Generate a random location $X^{\langle i,0 \rangle}$ using Equation \ref{equation1}\;
}

\For{$t \gets 1$ \KwTo $Iter_{\max}$}{
    \For{$i\gets 1$ \KwTo $N$}{
        $\theta_0 \leftarrow $ generate a random perceiving angle using Equation \ref{equation6}\;

        \tcc{Shooting behavior (exploration of the search space)}
        \If{($|\theta_0| \in ]0,\theta[\cup]\pi-\theta,\pi[$)}{
            Compute $X^{\langle i,t \rangle}_{prey}$ using Equation \ref{equation3}\;
            \For{$j\gets 1$ \KwTo $N$}{
                \If{($f(X^{\langle i,t \rangle}_{prey}) < f(X^{\langle j,t \rangle}) $)}{
                    Update the location $X^{\langle j,t \rangle}$ using Equation \ref{equation2}, and adjust its components\;
                }\Else{
                    If the location $X^{\langle j,t \rangle}$ has not been changed for a given number of iterations. In this case, generate a new location for $X^{\langle j,t \rangle}$ using Equations \ref{equation7} and \ref{equation8}, and adjust its components\;
                }
            }
        }

        \tcc{Jumping behavior (exploitation of the search space)}
        \Else{
            Compute $X^{\langle i,t \rangle}_{prey}$ using Equation \ref{equation5}\;
            \If{($f(X^{\langle i,t \rangle}_{prey}) < f(X^{\langle i,t \rangle}) $)}{
                Update the location $X^{\langle i,t \rangle}$ using Equation \ref{equation4}, and adjust its components\;
            }\Else{
                If the location $X^{\langle i,t \rangle}$ has not been changed for a given number of iterations. In this case, generate a new location for $X^{\langle i,t \rangle}$ using Equations \ref{equation7} and \ref{equation8}, and adjust its components\;
            }
        }
    }
}
\caption{The Archerfish Hunting Optimizer.}
\label{algorithm1}
\end{algorithm}

AHO's computational complexity depends on the following steps: initialization, fitness evaluation, and updating of candidate solutions. The computational complexity of the first step is $O(N)$. The computational complexity of the second step is $O(Iter_{\max} \times N \times d)$. The computational complexity of the third step is $O(Iter_{\max} \times N \times N)$  Therefore, the computational complexity of AHO is $O(N \times (Iter_{\max} \times (N + d)+1))$.

\section{Experimental results on unconstrained optimization problems}
\label{section4}

The benchmark CEC 2020 for unconstrained optimization problems \cite{yue2019problem} is used to investigate the effectiveness of AHO. This benchmark is composed of four groups of test functions: unimodal (UM), basic (BC), hybrid (HD), and composition (CM). The UM functions have one global optimum. They are used to assess the exploitation's capacity of AHO. The BC functions possess several global optimums. They are employed to evaluate, on the one hand, the exploration's ability of AHO, and on the other hand, its potential of local optimums' avoidance. The HD and CM functions are obtained from the hybridization and the composition of several elementary test functions. They are utilized to disclose the well-balancing between the exploration and exploitation of AHO. The mathematical formulation and characteristics of UM, BC, HD, and CM functions are available in \cite{yue2019problem}.

All the experiments were run using the Java programming language on a workstation with a Windows 10 familial edition (64-bit). The processor is Intel(R) Core(TM) i7--9750H CPU @ 2.60GHz 2.59 GHz, with 16 GB of RAM. The dimension of the search space ($d$) is set to 5, 10, 15, or 20. The population size ($N$) is set to $\lfloor 30 \times d^{1.5} \rfloor$, where the term $\lfloor x \rfloor$ expresses the truncation of the real number $x$. For each dimension, the maximum number of iterations ($Iter_{\max}$) is equal to $\frac{50000}{N}$, $\frac{1000000}{N}$, $\frac{3000000}{N}$, or $\frac{10000000}{N}$, respectively. The range of allowed values for each decision variable is $[-100,100]$. All the results are averaged over 30 independent runs. The swapping angle values ($\theta$) are set to $\frac{\pi}{12}$, $\frac{\pi}{6}$, $\frac{\pi}{4}$, $\frac{\pi}{3}$, or $\frac{5\pi}{12}$. The attractiveness rate values ($\omega$) are estimated to 0.01, 0.05, 0.25, 1.25, or 6.25. Hence, in all, we have 25 different configurations for each dimension. Tables \ref{table3}, \ref{table4}, \ref{table5}, and \ref{table6} present the standard deviation values (STD) for each dimension. It is worth mentioning that the closer the value of STD is to 0, the closer the result is to the global optimum.

We perform the Friedman test \cite{zimmerman1993relative} to show if the tuning of controlling parameters (i.e., $\theta$ and $\omega$) impacts the performance of AHO. We consider each dimension separately (see Tables \ref{table3}, \ref{table4}, \ref{table5}, and \ref{table6}). When $d=5$, we have 25 configurations (i.e., treatments) and 8 test functions (i.e., blocks). When $d=10$, $15$, or $20$, we have 25 configurations (i.e., treatments) and 10 test functions (i.e., blocks). The value of $\alpha$ is set to 0.05, and the value of the degree of freedom ($df$) is set to 24. We define the null and alternative hypotheses respectively as follows: $H_0$ there is no difference between the 25 configurations for each dimension, and $H_1$ there is a difference between the 25 configurations  for each dimension. The critical value for $\alpha=0.05$ and $df=24$ is 36.4150 \cite{2006586}. We compute the $F_r$ value using Equation \ref{equation9} \cite{zimmerman1993relative}, where $n$ is the number of blocks, $k$ is the number of treatments, and $T_1,T_2,\ldots,T_k$ are the sums of ranks for each treatment separately. If $F_r$ is greater than 36.4150, we reject the hypothesis $H_0$.

\begin{equation}
\label{equation9}
F_r=\frac{12}{nk(k+1)}(T_1^2+T_2^2+\ldots+T_k^2)-3n(k+1)
\end{equation}

\begin{itemize}
  \item For Table \ref{table3}, we have $F_r=66.7419 >36.4150$. Hence, the hypothesis $H_0$ is rejected. The best configuration is when $\theta=\frac{\pi}{12}$ and $\omega=0.01$ because its rank is the smallest with the value 57.5.
  \item For Table \ref{table4}, we have $F_r=104.2532 > 36.4150$. Hence, the hypothesis $H_0$ is rejected. The best configuration is when $\theta=\frac{5\pi}{12}$ and $\omega=0.01$ because its rank is the smallest with the value 68.
  \item For Table \ref{table5}, we have $F_r=125.8588 > 36.4150$. Hence, the null hypothesis $H_0$ is rejected. The best configuration is when $\theta=\frac{\pi}{3}$ and $\omega=0.01$ because its rank is the smallest with the value 58.
  \item For Table \ref{table6}, we have $F_r=144.8797 > 36.4150$. Hence, the null hypothesis $H_0$ is rejected. The best configuration is when $\theta=\frac{5\pi}{12}$ and $\omega=0.01$ because its rank is the smallest with the value 52.
\end{itemize}

\begin{table}[!]
\centering
\caption{The standard deviation values for dimension $d=5$.}
\label{table3}
\resizebox{\textwidth}{!}{
\begin{tabular}{llllllllll}
\hline
\multicolumn{2}{l}{} & $F_{1}$ & $F_{2}$ & $F_{3}$ & $F_{4}$ & $F_{5}$ & $F_{8}$ & $F_{9}$ & $F_{10}$ \\ \hline
\multirow{5}{*}{$\theta=\frac{\pi}{12}$} & $\omega=0.01$ & 0,00E+00 & 6,36E-05 & 0,00E+00 & 0,00E+00 & 1,94E-05 & 5,59E-05 & 2,04E-08 & 1,46E-03\\
 & $\omega=0.05$ & 0,00E+00 & 6,36E-05 & 0,00E+00 & 0,00E+00 & 2,77E-05 & 5,59E-05 & 1,33E-07 & 1,96E-03\\
 & $\omega=0.25$ & 1,78E-07 & 6,36E-05 & 0,00E+00 & 0,00E+00 & 5,20E-05 & 5,59E-05 & 4,28E-07 & 3,42E-03\\
 & $\omega=1.25$ & 1,52E-06 & 6,36E-05 & 3,12E-08 & 0,00E+00 & 9,38E-05 & 5,59E-05 & 3,66E-06 & 3,16E-03\\
 & $\omega=6.25$ & 1,16E-05 & 6,36E-05 & 1,58E-06 & 0,00E+00 & 1,66E-04 & 5,59E-05 & 1,37E-05 & 4,35E-03\\ \hline
\multirow{5}{*}{$\theta=\frac{\pi}{6}$} & $\omega=0.01$ & 0,00E+00 & 6,36E-05 & 0,00E+00 & 0,00E+00 & 2,14E-05 & 5,59E-05 & 1,75E-08 & 1,80E-03\\
 & $\omega=0.05$ & 1,41E-08 & 6,36E-05 & 0,00E+00 & 0,00E+00 & 3,27E-05 & 5,59E-05 & 9,89E-08 & 2,51E-03\\
 & $\omega=0.25$ & 1,44E-07 & 6,36E-05 & 0,00E+00 & 0,00E+00 & 4,83E-05 & 5,59E-05 & 3,54E-07 & 2,57E-03\\
 & $\omega=1.25$ & 1,43E-06 & 6,36E-05 & 3,02E-08 & 0,00E+00 & 9,55E-05 & 5,59E-05 & 3,71E-06 & 4,11E-03\\
 & $\omega=6.25$ & 1,01E-05 & 6,36E-05 & 1,06E-06 & 0,00E+00 & 1,61E-04 & 5,59E-05 & 5,56E-06 & 4,22E-03\\ \hline
\multirow{5}{*}{$\theta=\frac{\pi}{4}$} & $\omega=0.01$ & 0,00E+00 & 6,36E-05 & 0,00E+00 & 0,00E+00 & 1,83E-05 & 5,59E-05 & 2,26E-08 & 1,81E-03\\
 & $\omega=0.05$ & 1,58E-08 & 6,36E-05 & 0,00E+00 & 0,00E+00 & 2,84E-05 & 5,59E-05 & 9,87E-08 & 2,15E-03\\
 & $\omega=0.25$ & 8,39E-08 & 6,36E-05 & 0,00E+00 & 0,00E+00 & 4,28E-05 & 5,59E-05 & 2,78E-07 & 3,02E-03\\
 & $\omega=1.25$ & 1,44E-06 & 6,36E-05 & 4,33E-08 & 0,00E+00 & 1,02E-04 & 5,59E-05 & 4,43E-06 & 3,57E-03\\
 & $\omega=6.25$ & 1,41E-05 & 6,36E-05 & 6,71E-07 & 0,00E+00 & 1,98E-04 & 5,59E-05 & 9,97E-06 & 4,08E-03\\ \hline
\multirow{5}{*}{$\theta=\frac{\pi}{3}$} & $\omega=0.01$ & 0,00E+00 & 6,36E-05 & 0,00E+00 & 0,00E+00 & 1,87E-05 & 5,59E-05 & 2,62E-08 & 1,64E-03\\
 & $\omega=0.05$ & 0,00E+00 & 6,36E-05 & 0,00E+00 & 0,00E+00 & 2,82E-05 & 5,59E-05 & 9,44E-08 & 2,57E-03\\
 & $\omega=0.25$ & 8,98E-08 & 6,36E-05 & 0,00E+00 & 0,00E+00 & 3,91E-05 & 5,59E-05 & 7,26E-07 & 2,60E-03\\
 & $\omega=1.25$ & 1,27E-06 & 6,36E-05 & 4,17E-08 & 0,00E+00 & 8,51E-05 & 5,59E-05 & 6,87E-07 & 3,32E-03\\
 & $\omega=6.25$ & 1,64E-05 & 6,36E-05 & 8,99E-07 & 0,00E+00 & 2,56E-04 & 5,59E-05 & 1,35E-05 & 4,48E-03\\ \hline
\multirow{5}{*}{$\theta=\frac{5\pi}{12}$} & $\omega=0.01$ & 0,00E+00 & 6,36E-05 & 0,00E+00 & 0,00E+00 & 2,00E-05 & 5,59E-05 & 2,33E-08 & 1,74E-03\\
 & $\omega=0.05$ & 0,00E+00 & 6,36E-05 & 0,00E+00 & 0,00E+00 & 2,39E-05 & 5,59E-05 & 9,41E-08 & 2,22E-03\\
 & $\omega=0.25$ & 8,15E-08 & 6,36E-05 & 0,00E+00 & 0,00E+00 & 4,69E-05 & 5,59E-05 & 4,27E-07 & 2,84E-03\\
 & $\omega=1.25$ & 4,69E-07 & 6,36E-05 & 3,73E-08 & 0,00E+00 & 8,75E-05 & 5,59E-05 & 7,58E-07 & 3,14E-03\\
 & $\omega=6.25$ & 2,35E-05 & 6,36E-05 & 1,09E-06 & 0,00E+00 & 1,84E-04 & 5,59E-05 & 1,37E-05 & 4,39E-03\\ \hline
\end{tabular}
}
\end{table}

\begin{table}[!]
\centering
\caption{The standard deviation values for dimension $d=10$.}
\label{table4}
\resizebox{\textwidth}{!}{
\begin{tabular}{llllllllllll}
\hline
\multicolumn{2}{l}{} & $F_{1}$ & $F_{2}$ & $F_{3}$ & $F_{4}$ & $F_{5}$ & $F_{6}$ & $F_{7}$ & $F_{8}$ & $F_{9}$ & $F_{10}$ \\ \hline
\multirow{5}{*}{$\theta=\frac{\pi}{12}$} & $\omega=0.01$ & 4,55E-08 & 1,27E-04 & 0,00E+00 & 0,00E+00 & 3,82E-05 & 4,74E-05 & 4,44E-05 & 1,09E-04 & 0,00E+00 & 2,94E-03\\
 & $\omega=0.05$ & 2,71E-07 & 1,27E-04 & 0,00E+00 & 0,00E+00 & 3,82E-05 & 5,23E-05 & 5,18E-05 & 1,09E-04 & 0,00E+00 & 3,97E-03\\
 & $\omega=0.25$ & 7,83E-06 & 1,27E-04 & 5,77E-08 & 0,00E+00 & 3,82E-05 & 6,49E-05 & 8,36E-05 & 1,09E-04 & 0,00E+00 & 5,60E-03\\
 & $\omega=1.25$ & 2,08E-04 & 1,27E-04 & 2,24E-06 & 0,00E+00 & 3,87E-05 & 9,97E-05 & 1,43E-04 & 1,09E-04 & 2,80E-08 & 6,08E-03\\
 & $\omega=6.25$ & 5,55E-03 & 1,27E-04 & 4,68E-05 & 0,00E+00 & 4,02E-05 & 2,70E-04 & 5,38E-04 & 1,09E-04 & 7,02E-07 & 6,82E-03\\ \hline
\multirow{5}{*}{$\theta=\frac{\pi}{6}$} & $\omega=0.01$ & 4,30E-08 & 1,27E-04 & 0,00E+00 & 0,00E+00 & 3,82E-05 & 4,51E-05 & 3,95E-05 & 1,09E-04 & 0,00E+00 & 3,31E-03\\
 & $\omega=0.05$ & 4,96E-07 & 1,27E-04 & 0,00E+00 & 0,00E+00 & 3,83E-05 & 5,32E-05 & 5,49E-05 & 1,09E-04 & 0,00E+00 & 3,81E-03\\
 & $\omega=0.25$ & 7,04E-06 & 1,27E-04 & 5,16E-08 & 0,00E+00 & 3,84E-05 & 7,53E-05 & 9,26E-05 & 1,09E-04 & 0,00E+00 & 5,15E-03\\
 & $\omega=1.25$ & 2,11E-04 & 1,27E-04 & 2,19E-06 & 0,00E+00 & 3,89E-05 & 1,30E-04 & 1,81E-04 & 1,09E-04 & 2,76E-08 & 5,85E-03\\
 & $\omega=6.25$ & 3,79E-03 & 1,27E-04 & 4,97E-05 & 0,00E+00 & 3,98E-05 & 2,58E-04 & 5,94E-04 & 1,09E-04 & 7,54E-07 & 8,15E-03\\ \hline
\multirow{5}{*}{$\theta=\frac{\pi}{4}$} & $\omega=0.01$ & 3,68E-08 & 1,27E-04 & 0,00E+00 & 0,00E+00 & 3,82E-05 & 4,54E-05 & 3,63E-05 & 1,09E-04 & 0,00E+00 & 3,47E-03\\
 & $\omega=0.05$ & 5,90E-07 & 1,27E-04 & 0,00E+00 & 0,00E+00 & 3,82E-05 & 5,49E-05 & 5,30E-05 & 1,09E-04 & 0,00E+00 & 3,58E-03\\
 & $\omega=0.25$ & 9,16E-06 & 1,27E-04 & 5,93E-08 & 0,00E+00 & 3,84E-05 & 8,31E-05 & 8,57E-05 & 1,09E-04 & 0,00E+00 & 5,46E-03\\
 & $\omega=1.25$ & 2,01E-04 & 1,27E-04 & 1,83E-06 & 0,00E+00 & 3,87E-05 & 1,36E-04 & 2,09E-04 & 1,09E-04 & 3,11E-08 & 7,60E-03\\
 & $\omega=6.25$ & 4,04E-03 & 1,27E-04 & 3,70E-05 & 0,00E+00 & 4,06E-05 & 2,46E-04 & 5,57E-04 & 1,09E-04 & 7,41E-07 & 7,55E-03\\ \hline
\multirow{5}{*}{$\theta=\frac{\pi}{3}$} & $\omega=0.01$ & 3,53E-08 & 1,27E-04 & 0,00E+00 & 0,00E+00 & 3,82E-05 & 4,46E-05 & 4,30E-05 & 1,09E-04 & 0,00E+00 & 3,57E-03\\
 & $\omega=0.05$ & 4,63E-07 & 1,27E-04 & 0,00E+00 & 0,00E+00 & 3,82E-05 & 5,68E-05 & 6,02E-05 & 1,09E-04 & 0,00E+00 & 3,79E-03\\
 & $\omega=0.25$ & 1,04E-05 & 1,27E-04 & 8,70E-08 & 0,00E+00 & 3,84E-05 & 7,60E-05 & 1,04E-04 & 1,09E-04 & 0,00E+00 & 6,02E-03\\
 & $\omega=1.25$ & 2,30E-04 & 1,27E-04 & 1,88E-06 & 0,00E+00 & 3,94E-05 & 1,24E-04 & 1,71E-04 & 1,09E-04 & 2,41E-08 & 6,68E-03\\
 & $\omega=6.25$ & 5,94E-03 & 1,27E-04 & 5,31E-05 & 0,00E+00 & 3,99E-05 & 2,49E-04 & 4,32E-04 & 1,09E-04 & 7,23E-07 & 8,14E-03\\ \hline
\multirow{5}{*}{$\theta=\frac{5\pi}{12}$} & $\omega=0.01$ & 1,36E-08 & 1,27E-04 & 0,00E+00 & 0,00E+00 & 3,82E-05 & 4,58E-05 & 3,83E-05 & 1,09E-04 & 0,00E+00 & 3,37E-03\\
 & $\omega=0.05$ & 4,78E-07 & 1,27E-04 & 0,00E+00 & 0,00E+00 & 3,82E-05 & 5,23E-05 & 5,80E-05 & 1,09E-04 & 0,00E+00 & 4,61E-03\\
 & $\omega=0.25$ & 1,10E-05 & 1,27E-04 & 5,59E-08 & 0,00E+00 & 3,84E-05 & 7,95E-05 & 9,62E-05 & 1,09E-04 & 0,00E+00 & 4,44E-03\\
 & $\omega=1.25$ & 1,81E-04 & 1,27E-04 & 2,10E-06 & 0,00E+00 & 3,86E-05 & 1,40E-04 & 1,63E-04 & 1,09E-04 & 2,93E-08 & 5,76E-03\\
 & $\omega=6.25$ & 2,59E-03 & 1,27E-04 & 5,23E-05 & 0,00E+00 & 4,13E-05 & 2,55E-04 & 5,91E-04 & 1,09E-04 & 7,24E-07 & 8,28E-03\\ \hline
 \end{tabular}
}
\end{table}

\begin{table}[!]
\centering
\caption{The standard deviation values for dimension $d=15$.}
\label{table5}
\resizebox{\textwidth}{!}{
\begin{tabular}{llllllllllll}
\hline
\multicolumn{2}{l}{} & $F_{1}$ & $F_{2}$ & $F_{3}$ & $F_{4}$ & $F_{5}$ & $F_{6}$ & $F_{7}$ & $F_{8}$ & $F_{9}$ & $F_{10}$ \\ \hline
\multirow{5}{*}{$\theta=\frac{\pi}{12}$} & $\omega=0.01$ & 1,75E-07 & 1,91E-04 & 0,00E+00 & 0,00E+00 & 6,36E-05 & 5,61E-05 & 5,64E-05 & 1,52E-04 & 0,00E+00 & 8,56E-03\\
 & $\omega=0.05$& 3,46E-06 & 1,91E-04 & 1,64E-08 & 0,00E+00 & 6,37E-05 & 6,47E-05 & 7,13E-05 & 1,52E-04 & 0,00E+00 & 1,23E-02\\
 & $\omega=0.25$& 5,06E-05 & 1,91E-04 & 4,61E-07 & 0,00E+00 & 6,37E-05 & 8,23E-05 & 1,34E-04 & 1,52E-04 & 1,35E-08 & 1,90E-02\\
 & $\omega=1.25$& 1,15E-03 & 1,91E-04 & 1,14E-05 & 0,00E+00 & 6,43E-05 & 1,33E-04 & 2,79E-04 & 1,52E-04 & 3,17E-07 & 2,56E-02\\
 & $\omega=6.25$& 2,93E-02 & 1,91E-04 & 2,31E-04 & 0,00E+00 & 8,48E-05 & 2,96E-04 & 8,09E-04 & 1,52E-04 & 6,83E-06 & 2,78E-02\\ \hline
\multirow{5}{*}{$\theta=\frac{\pi}{6}$} & $\omega=0.01$ & 1,20E-07 & 1,91E-04 & 0,00E+00 & 0,00E+00 & 6,36E-05 & 5,48E-05 & 5,81E-05 & 1,52E-04 & 0,00E+00 & 9,22E-03\\
 & $\omega=0.05$& 2,77E-06 & 1,91E-04 & 1,20E-08 & 0,00E+00 & 6,36E-05 & 6,67E-05 & 7,26E-05 & 1,52E-04 & 0,00E+00 & 1,37E-02\\
 & $\omega=0.25$& 6,11E-05 & 1,91E-04 & 4,25E-07 & 0,00E+00 & 6,38E-05 & 9,27E-05 & 1,26E-04 & 1,52E-04 & 1,26E-08 & 1,94E-02\\
 & $\omega=1.25$& 1,23E-03 & 1,91E-04 & 1,34E-05 & 0,00E+00 & 6,59E-05 & 1,28E-04 & 2,08E-04 & 1,52E-04 & 4,28E-07 & 2,53E-02\\
 & $\omega=6.25$& 2,50E-02 & 1,92E-04 & 2,17E-04 & 1,14E-08 & 7,15E-05 & 2,83E-04 & 9,32E-04 & 1,52E-04 & 6,86E-06 & 3,11E-02\\ \hline
\multirow{5}{*}{$\theta=\frac{\pi}{4}$} & $\omega=0.01$ & 1,37E-07 & 1,91E-04 & 0,00E+00 & 0,00E+00 & 6,36E-05 & 5,80E-05 & 5,64E-05 & 1,52E-04 & 0,00E+00 & 8,59E-03\\
 & $\omega=0.05$& 2,66E-06 & 1,91E-04 & 1,55E-08 & 0,00E+00 & 6,37E-05 & 6,62E-05 & 7,61E-05 & 1,52E-04 & 0,00E+00 & 1,33E-02\\
 & $\omega=0.25$& 4,35E-05 & 1,91E-04 & 3,65E-07 & 0,00E+00 & 6,38E-05 & 9,05E-05 & 1,13E-04 & 1,52E-04 & 1,24E-08 & 1,52E-02\\
 & $\omega=1.25$& 1,01E-03 & 1,91E-04 & 1,13E-05 & 0,00E+00 & 6,49E-05 & 1,53E-04 & 2,20E-04 & 1,52E-04 & 3,01E-07 & 2,14E-02\\
 & $\omega=6.25$& 2,69E-02 & 1,92E-04 & 3,55E-04 & 0,00E+00 & 7,49E-05 & 2,60E-04 & 7,91E-04 & 1,52E-04 & 8,24E-06 & 2,81E-02\\ \hline
\multirow{5}{*}{$\theta=\frac{\pi}{3}$} & $\omega=0.01$ & 8,59E-08 & 1,91E-04 & 0,00E+00 & 0,00E+00 & 6,36E-05 & 5,64E-05 & 5,49E-05 & 1,52E-04 & 0,00E+00 & 9,07E-03\\
 & $\omega=0.05$& 2,14E-06 & 1,91E-04 & 1,53E-08 & 0,00E+00 & 6,37E-05 & 6,79E-05 & 6,76E-05 & 1,52E-04 & 0,00E+00 & 1,35E-02\\
 & $\omega=0.25$& 3,09E-05 & 1,91E-04 & 3,77E-07 & 0,00E+00 & 6,40E-05 & 8,43E-05 & 1,18E-04 & 1,52E-04 & 1,46E-08 & 1,67E-02\\
 & $\omega=1.25$& 9,38E-04 & 1,91E-04 & 1,02E-05 & 0,00E+00 & 6,64E-05 & 1,44E-04 & 2,41E-04 & 1,52E-04 & 3,34E-07 & 2,51E-02\\
 & $\omega=6.25$& 3,02E-02 & 1,91E-04 & 2,54E-04 & 1,48E-08 & 7,67E-05 & 3,16E-04 & 8,32E-04 & 1,52E-04 & 7,65E-06 & 2,70E-02\\ \hline
\multirow{5}{*}{$\theta=\frac{5\pi}{12}$} & $\omega=0.01$ & 1,39E-07 & 1,91E-04 & 0,00E+00 & 0,00E+00 & 6,36E-05 & 5,60E-05 & 5,54E-05 & 1,52E-04 & 0,00E+00 & 9,71E-03\\
 & $\omega=0.05$& 3,84E-06 & 1,91E-04 & 1,52E-08 & 0,00E+00 & 6,37E-05 & 6,15E-05 & 6,85E-05 & 1,52E-04 & 0,00E+00 & 1,35E-02\\
 & $\omega=0.25$& 6,00E-05 & 1,91E-04 & 3,43E-07 & 0,00E+00 & 6,39E-05 & 8,24E-05 & 1,14E-04 & 1,52E-04 & 1,42E-08 & 1,61E-02\\
 & $\omega=1.25$& 1,46E-03 & 1,91E-04 & 1,13E-05 & 0,00E+00 & 6,50E-05 & 1,58E-04 & 2,39E-04 & 1,52E-04 & 3,94E-07 & 2,26E-02\\
 & $\omega=6.25$& 2,12E-02 & 1,91E-04 & 2,66E-04 & 0,00E+00 & 7,21E-05 & 3,00E-04 & 5,61E-04 & 1,52E-04 & 1,05E-05 & 2,76E-02\\ \hline
\end{tabular}
}
\end{table}

\begin{table}[!]
\centering
\caption{The standard deviation values for dimension $d=20$.}
\label{table6}
\resizebox{\textwidth}{!}{
\begin{tabular}{llllllllllll}
\hline
\multicolumn{2}{l}{} & $F_{1}$ & $F_{2}$ & $F_{3}$ & $F_{4}$ & $F_{5}$ & $F_{6}$ & $F_{7}$ & $F_{8}$ & $F_{9}$ & $F_{10}$ \\ \hline
\multirow{5}{*}{$\theta=\frac{\pi}{12}$} & $\omega=0.01$ & 4,99E-07 & 2,55E-04 & 0,00E+00 & 0,00E+00 & 7,68E-05 & 1,02E-04 & 7,70E-05 & 2,00E-04 & 0,00E+00 & 9,84E-03\\
 & $\omega=0.05$& 5,51E-06 & 2,55E-04 & 7,33E-08 & 0,00E+00 & 7,76E-05 & 1,24E-04 & 9,12E-05 & 2,00E-04 & 0,00E+00 & 1,23E-02\\
 & $\omega=0.25$& 1,74E-04 & 2,55E-04 & 1,94E-06 & 0,00E+00 & 8,25E-05 & 1,68E-04 & 1,62E-04 & 2,00E-04 & 2,39E-08 & 1,48E-02\\
 & $\omega=1.25$& 3,85E-03 & 2,55E-04 & 4,56E-05 & 0,00E+00 & 9,08E-05 & 2,51E-04 & 3,23E-04 & 2,00E-04 & 7,21E-07 & 1,61E-02\\
 & $\omega=6.25$& 9,70E-02 & 2,56E-04 & 1,18E-03 & 5,45E-08 & 1,06E-04 & 8,19E-04 & 1,06E-03 & 2,00E-04 & 1,39E-05 & 2,20E-02\\ \hline
\multirow{5}{*}{$\theta=\frac{\pi}{6}$} & $\omega=0.01$ & 6,77E-07 & 2,55E-04 & 0,00E+00 & 0,00E+00 & 7,67E-05 & 1,03E-04 & 6,94E-05 & 2,00E-04 & 0,00E+00 & 9,34E-03\\
 & $\omega=0.05$ & 7,47E-06 & 2,55E-04 & 6,52E-08 & 0,00E+00 & 7,82E-05 & 1,14E-04 & 9,49E-05 & 2,00E-04 & 0,00E+00 & 1,09E-02\\
 & $\omega=0.25$ & 2,23E-04 & 2,55E-04 & 1,50E-06 & 0,00E+00 & 8,53E-05 & 1,94E-04 & 1,52E-04 & 2,00E-04 & 2,63E-08 & 1,48E-02\\
 & $\omega=1.25$ & 4,83E-03 & 2,55E-04 & 5,00E-05 & 0,00E+00 & 9,10E-05 & 3,65E-04 & 3,27E-04 & 2,00E-04 & 6,01E-07 & 1,64E-02\\
 & $\omega=6.25$ & 1,11E-01 & 2,56E-04 & 1,08E-03 & 3,13E-08 & 1,10E-04 & 7,06E-04 & 1,01E-03 & 2,00E-04 & 1,74E-05 & 2,21E-02\\ \hline
\multirow{5}{*}{$\theta=\frac{\pi}{4}$} & $\omega=0.01$ & 4,42E-07 & 2,55E-04 & 0,00E+00 & 0,00E+00 & 7,69E-05 & 1,02E-04 & 7,23E-05 & 2,00E-04 & 0,00E+00 & 9,39E-03\\
 & $\omega=0.05$ & 1,11E-05 & 2,55E-04 & 7,81E-08 & 0,00E+00 & 7,92E-05 & 1,20E-04 & 1,04E-04 & 2,00E-04 & 0,00E+00 & 1,06E-02\\
 & $\omega=0.25$ & 2,21E-04 & 2,55E-04 & 1,50E-06 & 0,00E+00 & 8,31E-05 & 1,73E-04 & 1,60E-04 & 2,00E-04 & 2,89E-08 & 1,54E-02\\
 & $\omega=1.25$ & 3,39E-03 & 2,55E-04 & 4,15E-05 & 0,00E+00 & 8,87E-05 & 3,09E-04 & 3,79E-04 & 2,00E-04 & 7,14E-07 & 1,56E-02\\
 & $\omega=6.25$ & 1,37E-01 & 2,55E-04 & 1,43E-03 & 3,15E-08 & 1,17E-04 & 7,20E-04 & 1,16E-03 & 2,00E-04 & 1,68E-05 & 2,05E-02\\ \hline
\multirow{5}{*}{$\theta=\frac{\pi}{3}$} & $\omega=0.01$ & 4,87E-07 & 2,55E-04 & 0,00E+00 & 0,00E+00 & 7,67E-05 & 1,04E-04 & 7,21E-05 & 2,00E-04 & 0,00E+00 & 9,62E-03\\
 & $\omega=0.05$ & 1,33E-05 & 2,55E-04 & 5,31E-08 & 0,00E+00 & 7,86E-05 & 1,19E-04 & 9,82E-05 & 2,00E-04 & 0,00E+00 & 1,05E-02\\
 & $\omega=0.25$ & 1,43E-04 & 2,55E-04 & 2,07E-06 & 0,00E+00 & 8,09E-05 & 1,85E-04 & 1,65E-04 & 2,00E-04 & 2,46E-08 & 1,38E-02\\
 & $\omega=1.25$ & 3,63E-03 & 2,55E-04 & 4,64E-05 & 0,00E+00 & 8,94E-05 & 3,54E-04 & 2,94E-04 & 2,00E-04 & 7,25E-07 & 1,70E-02\\
 & $\omega=6.25$ & 1,33E-01 & 2,56E-04 & 7,94E-04 & 2,71E-08 & 1,14E-04 & 7,15E-04 & 1,01E-03 & 2,00E-04 & 1,59E-05 & 2,17E-02\\ \hline
\multirow{5}{*}{$\theta=\frac{5\pi}{12}$} & $\omega=0.01$ & 5,09E-07 & 2,55E-04 & 0,00E+00 & 0,00E+00 & 7,67E-05 & 1,00E-04 & 6,74E-05 & 2,00E-04 & 0,00E+00 & 8,83E-03\\
 & $\omega=0.05$ & 1,06E-05 & 2,55E-04 & 6,13E-08 & 0,00E+00 & 7,91E-05 & 1,29E-04 & 1,07E-04 & 2,00E-04 & 0,00E+00 & 1,15E-02\\
 & $\omega=0.25$ & 1,92E-04 & 2,55E-04 & 1,77E-06 & 0,00E+00 & 8,25E-05 & 1,81E-04 & 1,69E-04 & 2,00E-04 & 3,08E-08 & 1,32E-02\\
 & $\omega=1.25$ & 3,39E-03 & 2,55E-04 & 5,49E-05 & 0,00E+00 & 8,97E-05 & 3,12E-04 & 2,61E-04 & 2,00E-04 & 7,85E-07 & 1,53E-02\\
 & $\omega=6.25$ & 1,06E-01 & 2,56E-04 & 1,28E-03 & 8,83E-08 & 1,24E-04 & 9,45E-04 & 1,05E-03 & 2,00E-04 & 1,54E-05 & 1,99E-02\\ \hline
\end{tabular}
}
\end{table}

From Tables \ref{table3}, \ref{table4}, \ref{table5}, and \ref{table6}, we observe that the problem's dimension does not significantly impact the quality of the obtained results. On the other hand, we notice that the perceiving angle and attractiveness rate's values do not influence the performance of AHO. In all configurations, we observe that AHO can expose excellent results. Its performance remains consistently the same when achieving runs with various decision variables and multiple values of its controlling parameters. We believe that such behavior is justified by the law of large numbers in probability theory \cite{etemadi1981elementary}. In other words, if the number of generations is large enough, AHO tends to reach an equitable balance between exploration and exploitation phases regardless of the value of perceiving angle.

AHO's statistical results on the benchmark CEC 2020 with 5, 10, 15, and 20 dimensions are summarized in Table \ref{table7}. It introduces the obtained best, worst, median, mean values, and the standard deviations of the error from the optimum solution over 30 independent runs for the ten benchmark functions. It is worth pointing out that the results presented in Table \ref{table7} are taken from the best configurations of Tables \ref{table3}, \ref{table4}, \ref{table5}, and \ref{table6} when performing the Friedman test.

\begin{table}[!]
\centering
\caption{Performance of AHO on 5d, 10d, 15d, and 20d functions averaged over 30 independent runs.}
\label{table7}
\resizebox{\textwidth}{!}{
\begin{tabular}{llllllllllll}
\hline
\multicolumn{2}{l}{} & $F_1$ & $F_2$ & $F_3$ & $F_4$ & $F_5$ & $F_6$ & $F_7$ & $F_8$ & $F_9$ & $F_{10}$ \\ \hline\hline
\multirow{5}{*}{\begin{tabular}[c]{@{}l@{}}$d=5$,\\$\theta=\frac{\pi}{12}$,\\and $\omega=0.01$\end{tabular}} & Worst & 8.54E-08 & 6.36E-05 & 0.00E+00 & 0.00E+00 & 3.70E-05 & - & - & 5.59E-05 & 4.58E-08 & 2.95E-03\\
 & Best& 0.00E+00 & 6.36E-05 & 0.00E+00 & 0.00E+00 & 1.48E-05 & - & - & 5.59E-05 & 0.00E+00 & 7.61E-04\\
 & Median& 0.00E+00 & 6.36E-05 & 0.00E+00 & 0.00E+00 & 2.27E-05 & - & - & 5.59E-05 & 2.62E-08 & 1.71E-03\\
 & Mean& 0.00E+00 & 6.36E-05 & 0.00E+00 & 0.00E+00 & 1.94E-05 & - & - & 5.59E-05 & 2.04E-08 & 1.46E-03\\
 & Std& 0.00E+00 & 6.36E-05 & 0.00E+00 & 0.00E+00 & 1.94E-05 & - & - & 5.59E-05 & 2.04E-08 & 1.46E-03\\ \hline
\multirow{5}{*}{\begin{tabular}[c]{@{}l@{}}$d=10$,\\$\theta=\frac{5\pi}{12}$,\\and $\omega=0.01$\end{tabular}} & Worst & 7.12E-07 & 1.27E-04 & 0.00E+00 & 0.00E+00 & 3.85E-05 & 6.59E-05 & 6.15E-05 & 1.09E-04 & 0.00E+00 & 5.67E-03\\
 & Best& 1.32E-08 & 1.27E-04 & 0.00E+00 & 0.00E+00 & 3.82E-05 & 4.13E-05 & 2.95E-05 & 1.09E-04 & 0.00E+00 & 2.04E-03\\
 & Median& 1.33E-07 & 1.27E-04 & 0.00E+00 & 0.00E+00 & 3.82E-05 & 5.05E-05 & 4.40E-05 & 1.09E-04 & 0.00E+00 & 3.64E-03\\
 & Mean& 1.36E-08 & 1.27E-04 & 0.00E+00 & 0.00E+00 & 3.82E-05 & 4.58E-05 & 3.83E-05 & 1.09E-04 & 0.00E+00 & 3.37E-03\\
 & Std& 1.36E-08 & 1.27E-04 & 0.00E+00 & 0.00E+00 & 3.82E-05 & 4.58E-05 & 3.83E-05 & 1.09E-04 & 0.00E+00 & 3.37E-03\\ \hline
\multirow{5}{*}{\begin{tabular}[c]{@{}l@{}}$d=15$,\\$\theta=\frac{\pi}{3}$,\\and $\omega=0.01$\end{tabular}} & Worst & 8.70E-07 & 1.91E-04 & 0.00E+00 & 0.00E+00 & 6.37E-05 & 6.62E-05 & 7.69E-05 & 1.52E-04 & 0.00E+00 & 1.34E-02\\
 & Best& 8.20E-08 & 1.91E-04 & 0.00E+00 & 0.00E+00 & 6.36E-05 & 5.37E-05 & 5.07E-05 & 1.52E-04 & 0.00E+00 & 7.33E-03\\
 & Median& 3.05E-07 & 1.91E-04 & 0.00E+00 & 0.00E+00 & 6.36E-05 & 5.88E-05 & 6.12E-05 & 1.52E-04 & 0.00E+00 & 9.74E-03\\
 & Mean& 8.59E-08 & 1.91E-04 & 0.00E+00 & 0.00E+00 & 6.36E-05 & 5.64E-05 & 5.49E-05 & 1.52E-04 & 0.00E+00 & 9.07E-03\\
 & Std& 8.59E-08 & 1.91E-04 & 0.00E+00 & 0.00E+00 & 6.36E-05 & 5.64E-05 & 5.49E-05 & 1.52E-04 & 0.00E+00 & 9.07E-03\\ \hline
\multirow{5}{*}{\begin{tabular}[c]{@{}l@{}}$d=20$,\\$\theta=\frac{5\pi}{12}$,\\and $\omega=0.01$\end{tabular}} & Worst & 4.91E-06 & 2.55E-04 & 1.20E-08 & 0.00E+00 & 7.81E-05 & 1.34E-04 & 1.05E-04 & 2.00E-04 & 0.00E+00 & 1.40E-02\\
 & Best& 4.16E-07 & 2.55E-04 & 0.00E+00 & 0.00E+00 & 7.65E-05 & 9.20E-05 & 5.96E-05 & 2.00E-04 & 0.00E+00 & 6.20E-03\\
 & Median& 1.30E-06 & 2.55E-04 & 0.00E+00 & 0.00E+00 & 7.70E-05 & 1.14E-04 & 8.00E-05 & 2.00E-04 & 0.00E+00 & 9.90E-03\\
 & Mean& 5.09E-07 & 2.55E-04 & 0.00E+00 & 0.00E+00 & 7.67E-05 & 1.00E-04 & 6.74E-05 & 2.00E-04 & 0.00E+00 & 8.83E-03\\
 & Std& 5.09E-07 & 2.55E-04 & 0.00E+00 & 0.00E+00 & 7.67E-05 & 1.00E-04 & 6.74E-05 & 2.00E-04 & 0.00E+00 & 8.83E-03\\ \hline
\end{tabular}
}
\end{table}

For function $F_{1}$, AHO successfully obtained the optimal solution in the dimension $d=5$ and was very close to the optimum in dimensions $d=10$, $d=15$, and $d=20$. For functions $F_{2}$, $F_{5}$, $F_{6}$, $F_{7}$, $F_{8}$, and $F_{10}$, AHO was very close to the optimum in all dimensions. For functions $F_{3}$ and $F_{4}$, AHO successfully reached the optimal solution in all dimensions. For function $F_{9}$, AHO successfully attained the optimal solution in dimensions $d=10$, $d=15$, and $d=20$ and was very close to the optimum in dimension $d=5$. In conclusion, we can say that:
\begin{itemize}
  \item From function $F_1$, AHO has a good exploitation of the search space.
  \item From functions $F_2$, $F_3$, and $F_{4}$, AHO has a good exploration of the search space and avoids getting trapped in local optimums.
  \item From functions $F_5$, $F_6$, $F_7$, $F_8$, $F_9$, and $F_{10}$, AHO has a good balance between the exploration and exploitation phases.
\end{itemize}

The performance of the archerfish hunting optimizer is compared to some well established metaheuristics algorithms mentioned next. The numerical results are reported in: i) Tables \ref{table8}, \ref{table9}, \ref{table10}, and \ref{table11} for $d=5$; ii) Tables \ref{table12}, \ref{table13}, \ref{table14}, and \ref{table15} for $d=10$, iii) Tables \ref{table16}, \ref{table17}, \ref{table18}, and \ref{table19} for $d=15$; and iv) Tables \ref{table20}, \ref{table21}, \ref{table22}, and \ref{table23} for $d=20$.

\begin{itemize}
  \item Improving Cuckoo Search: Incorporating Changes for CEC 2017 and CEC 2020 Benchmark Problems (CSsin) \cite{salgotra2020improving}.
  \item A Multi-Population Exploration-only Exploitation-only Hybrid on CEC-2020 Single Objective Bound Constrained Problems (MP-EEH) \cite{bolufe2020multi}.
  \item Ranked Archive Differential Evolution with Selective Pressure for CEC 2020 Numerical Optimization (RASPSHADE) \cite{stanovov2020ranked}.
  \item Improved Multi-operator Differential Evolution Algorithm for Solving Unconstrained Problems (IMODE) \cite{sallam2020improved}.
  \item DISH–XX Solving CEC2020 Single Objective Bound Constrained Numerical Optimization Benchmark (DISH-XX) \cite{viktorin2020dish}.
  \item Evaluating the Performance of Adaptive Gaining-Sharing Knowledge Based Algorithm on CEC 2020 Benchmark Problems (AGSK) \cite{mohamed2020evaluating}.
  \item Differential Evolution Algorithm for Single Objective Bound-Constrained Optimization: Algorithm j2020 (j2020) \cite{brest2020differential}.
  \item Eigenvector Crossover in jDE100 Algorithm (jDE100e) \cite{bujok2020eigenvector}.
  \item Large Initial Population and Neighborhood Search incoporated in LSHADE to solve CEC2020 Benchmark Problems (OLSHADE) \cite{biswas2020large}.
  \item Multi-population Modified L-SHADE for Single Objective Bound Constrained Optimization (mpmLSHADE) \cite{jou2020multi}.
  \item SOMA-CL for Competition on Single Objective Bound Constrained Numerical Optimization Benchmark (SOMA-CL) \cite{kadavy2020soma}.
  \item Gaining-sharing knowledge based algorithm for solving optimization problems: a novel nature inspired algorithm (GSK) \cite{mohamed2019gaining}.
\end{itemize}

We perform the Wilcoxon signed-rank test \cite{rey2011wilcoxon} to establish if AHO is better/worst than CSsin, MP-EEH, RASPSHADE, IMODE, DISH-XX, AGSK, j2020, jDE100e, OLSHADE, mpmLSHADE, SOMA-CL, and GSK in all dimensions. We consider the one-sided test with $\alpha=0.05$. For each dimension, the used values for controlling parameters of AHO are reported in Table \ref{table7}. For the algorithms selected for the comparative study, the parameters' values are the same as the recommended settings in their original works. Tables \ref{table24}, \ref{table25}, \ref{table26}, and \ref{table27} summarize the results of the Wilcoxon signed-rank test.

\begin{table}[!]
\centering
\caption{Statistical results of AHO in comparison to CSsin, MP-EEH, and RASPSHADE ($d=5$).}
\label{table8}
\resizebox{\textwidth}{!}{
\begin{tabular}{llllllllllll}
\hline
\multirow{2}{*}{} &
  \multicolumn{2}{c}{\textbf{AHO}} &
  \multicolumn{3}{c}{\textbf{CSsin}} &
  \multicolumn{3}{c}{\textbf{MP-EEH}} &
  \multicolumn{3}{c}{\textbf{RASPSHADE}} \\ \cline{2-12}
 &
  \multicolumn{1}{c}{\textbf{Mean}} &
  \multicolumn{1}{c}{\textbf{STD}} &
  \multicolumn{1}{c}{\textbf{Mean}} &
  \multicolumn{1}{c}{\textbf{STD}} &
  \multicolumn{1}{c}{\textbf{P}} &
  \multicolumn{1}{c}{\textbf{Mean}} &
  \multicolumn{1}{c}{\textbf{STD}} &
  \multicolumn{1}{c}{\textbf{P}} &
  \multicolumn{1}{c}{\textbf{Mean}} &
  \multicolumn{1}{c}{\textbf{STD}} &
  \multicolumn{1}{c}{\textbf{P}} \\ \hline
$F_{1}$ & 0,00E+00 & 0,00E+00 & 2,81E-07 & 3,01E-07 & + & 0,00E+00 & 0,00E+00 &=& 0,00E+00 & 0,00E+00 & =\\
$F_{2}$ & 6,36E-05 & 6,36E-05 & 8,75E+00 & 2,81E+01 & + & 2,46E+01 & 3,92E+01 & + & 5,64E-01 & 1,18E+00 & +\\
$F_{3}$ & 0,00E+00 & 0,00E+00 & 5,65E+00 & 4,74E-01 & + & 3,43E+00 & 2,13E+00 & + & 5,34E+00 & 1,72E-01 & +\\
$F_{4}$ & 0,00E+00 & 0,00E+00 & 0,00E+00 & 0,00E+00 &=& 1,02E-01 & 8,06E-02 & + & 1,04E-01 & 3,85E-02 & +\\
$F_{5}$ & 1,94E-05 & 1,94E-05 & 3,94E-04 & 6,52E-04 & + & 1,03E+01 & 9,90E+00 & + & 8,32E-02 & 2,12E-01 & +\\
$F_{8}$ & 5,59E-05 & 5,59E-05 & 3,67E+01 & 4,77E+01 & + & 1,68E-08 & 5,30E-09 & - & 0,00E+00 & 0,00E+00 & -\\
$F_{9}$ & 2,04E-08 & 2,04E-08 & 1,17E+01 & 3,25E+01 & + & 1,03E+02 & 1,83E+01 & + & 1,00E+02 & 0,00E+00 & -\\
$F_{10}$ & 1,46E-03 & 1,46E-03 & 2,88E+02 & 7,00E+01 & + & 3,10E+02 & 5,42E+01 & + & 3,44E+02 & 1,18E+01 & +\\ \hline
\end{tabular}
}
\end{table}

\begin{table}[!]
\centering
\caption{Statistical results of AHO in comparison to IMODE, DISH-XX, and AGSK ($d=5$).}
\label{table9}
\resizebox{\textwidth}{!}{
\begin{tabular}{llllllllllll}
\hline
\multirow{2}{*}{} &
  \multicolumn{2}{c}{\textbf{AHO}} &
  \multicolumn{3}{c}{\textbf{IMODE}} &
  \multicolumn{3}{c}{\textbf{DISH-XX}} &
  \multicolumn{3}{c}{\textbf{AGSK}} \\ \cline{2-12}
 &
  \multicolumn{1}{c}{\textbf{Mean}} &
  \multicolumn{1}{c}{\textbf{STD}} &
  \multicolumn{1}{c}{\textbf{Mean}} &
  \multicolumn{1}{c}{\textbf{STD}} &
  \multicolumn{1}{c}{\textbf{P}} &
  \multicolumn{1}{c}{\textbf{Mean}} &
  \multicolumn{1}{c}{\textbf{STD}} &
  \multicolumn{1}{c}{\textbf{P}} &
  \multicolumn{1}{c}{\textbf{Mean}} &
  \multicolumn{1}{c}{\textbf{STD}} &
  \multicolumn{1}{c}{\textbf{P}} \\ \hline
$F_{1}$ & 0,00E+00 & 0,00E+00 & 0,00E+00 & 0,00E+00 &=& 0,00E+00 & 0,00E+00 &=& 0,00E+00 & 0,00E+00 & =\\
$F_{2}$ & 6,36E-05 & 6,36E-05 & 8,33E-02 & 8,89E-02 & + & 1,31E+01 & 2,82E+01 & + & 1,64E+01 & 2,58E+01 & +\\
$F_{3}$ & 0,00E+00 & 0,00E+00 & 5,15E+00 & 0,00E+00 &=& 5,65E+00 & 6,07E-01 & + & 2,87E+00 & 2,05E+00 & +\\
$F_{4}$ & 0,00E+00 & 0,00E+00 & 0,00E+00 & 0,00E+00 &=& 1,42E-02 & 9,57E-03 & + & 1,11E-01 & 6,05E-02 & +\\
$F_{5}$ & 1,94E-05 & 1,94E-05 & 0,00E+00 & 0,00E+00 & - & 2,08E-01 & 2,99E-01 & + & 0,00E+00 & 0,00E+00 & -\\
$F_{8}$ & 5,59E-05 & 5,59E-05 & 0,00E+00 & 0,00E+00 & - & 3,35E+00 & 1,83E+01 & + & 0,00E+00 & 0,00E+00 & -\\
$F_{9}$ & 2,04E-08 & 2,04E-08 & 0,00E+00 & 0,00E+00 & - & 1,10E+02 & 4,03E+01 & + & 3,33E+01 & 4,79E+01 & +\\
$F_{10}$ & 1,46E-03 & 1,46E-03 & 2,44E+02 & 1,36E+02 & + & 3,44E+02 & 1,20E+01 & + & 2,25E+02 & 1,32E+02 & +\\ \hline
\end{tabular}
}
\end{table}

\begin{table}[!]
\centering
\caption{Statistical results of AHO in comparison to j2020, jDE100e, and OLSHADE ($d=5$).}
\label{table10}
\resizebox{\textwidth}{!}{
\begin{tabular}{llllllllllll}
\hline
\multirow{2}{*}{} &
  \multicolumn{2}{c}{\textbf{AHO}} &
  \multicolumn{3}{c}{\textbf{j2020}} &
  \multicolumn{3}{c}{\textbf{jDE100e}} &
  \multicolumn{3}{c}{\textbf{OLSHADE}} \\ \cline{2-12}
 &
  \multicolumn{1}{c}{\textbf{Mean}} &
  \multicolumn{1}{c}{\textbf{STD}} &
  \multicolumn{1}{c}{\textbf{Mean}} &
  \multicolumn{1}{c}{\textbf{STD}} &
  \multicolumn{1}{c}{\textbf{P}} &
  \multicolumn{1}{c}{\textbf{Mean}} &
  \multicolumn{1}{c}{\textbf{STD}} &
  \multicolumn{1}{c}{\textbf{P}} &
  \multicolumn{1}{c}{\textbf{Mean}} &
  \multicolumn{1}{c}{\textbf{STD}} &
  \multicolumn{1}{c}{\textbf{P}} \\ \hline
$F_{1}$ & 0,00E+00 & 0,00E+00 & 0,00E+00 & 0,00E+00 &=& 0,00E+00 & 0,00E+00 &=& 0,00E+00 & 0,00E+00 & =\\
$F_{2}$ & 6,36E-05 & 6,36E-05 & 3,23E+00 & 3,74E+00 & + & 1,34E+00 & 2,93E+00 & + & 1,88E+01 & 4,16E+01 & +\\
$F_{3}$ & 0,00E+00 & 0,00E+00 & 3,42E+00 & 2,33E+00 & + & 4,52E+00 & 1,52E+00 & + & 1,41E+00 & 1,36E+00 & +\\
$F_{4}$ & 0,00E+00 & 0,00E+00 & 7,68E-02 & 6,40E-02 & + & 8,67E-02 & 4,67E-02 & + & 7,27E-02 & 5,98E-02 & +\\
$F_{5}$ & 1,94E-05 & 1,94E-05 & 1,37E-01 & 2,86E-01 & + & 5,40E-02 & 2,11E-01 & + & 4,16E-02 & 1,58E-01 & +\\
$F_{8}$ & 5,59E-05 & 5,59E-05 & 6,28E-01 & 2,39E+00 & + & 1,20E-01 & 6,54E-01 & + & 0,00E+00 & 0,00E+00 & -\\
$F_{9}$ & 2,04E-08 & 2,04E-08 & 2,05E+01 & 3,75E+01 & + & 9,33E+01 & 2,54E+01 & + & 0,00E+00 & 0,00E+00 & -\\
$F_{10}$ & 1,46E-03 & 1,46E-03 & 1,26E+02 & 9,03E+01 & + & 2,90E+02 & 8,06E+01 & + & 1,13E+02 & 9,73E+01 & +\\ \hline
\end{tabular}
}
\end{table}

\begin{table}[!]
\centering
\caption{Statistical results of AHO in comparison to mpmLSHADE, SOMA-CL, and GSK ($d=5$).}
\label{table11}
\resizebox{\textwidth}{!}{
\begin{tabular}{llllllllllll}
\hline
\multirow{2}{*}{} &
  \multicolumn{2}{c}{\textbf{AHO}} &
  \multicolumn{3}{c}{\textbf{mpmLSHADE}} &
  \multicolumn{3}{c}{\textbf{SOMA-CL}} &
  \multicolumn{3}{c}{\textbf{GSK}} \\ \cline{2-12}
 &
  \multicolumn{1}{c}{\textbf{Mean}} &
  \multicolumn{1}{c}{\textbf{STD}} &
  \multicolumn{1}{c}{\textbf{Mean}} &
  \multicolumn{1}{c}{\textbf{STD}} &
  \multicolumn{1}{c}{\textbf{P}} &
  \multicolumn{1}{c}{\textbf{Mean}} &
  \multicolumn{1}{c}{\textbf{STD}} &
  \multicolumn{1}{c}{\textbf{P}} &
  \multicolumn{1}{c}{\textbf{Mean}} &
  \multicolumn{1}{c}{\textbf{STD}} &
  \multicolumn{1}{c}{\textbf{P}} \\ \hline
$F_{1}$ & 0,00E+00 & 0,00E+00 & 0,00E+00 & 0,00E+00 &=& 5,29E-02 & 3,35E-02 & + & 0,00E+00 & 0,00E+00 & =\\
$F_{2}$ & 6,36E-05 & 6,36E-05 & 1,17E-01 & 1,37E-01 & + & 6,67E+00 & 4,65E+00 & + & 1,23E+02 & 4,33E+01 & +\\
$F_{3}$ & 0,00E+00 & 0,00E+00 & 4,70E+00 & 1,47E+00 & + & 5,74E+00 & 4,29E-01 & + & 7,44E+00 & 6,40E-01 & +\\
$F_{4}$ & 0,00E+00 & 0,00E+00 & 9,61E-02 & 3,09E-02 & + & 2,18E-01 & 9,49E-02 & + & 3,53E-01 & 9,91E-02 & +\\
$F_{5}$ & 1,94E-05 & 1,94E-05 & 0,00E+00 & 0,00E+00 & - & 7,62E-03 & 1,06E-02 & + & 2,92E-02 & 6,49E-02 & +\\
$F_{8}$ & 5,59E-05 & 5,59E-05 & 3,33E+00 & 1,80E+01 & + & 2,02E-01 & 2,96E-01 & + & 6,78E-02 & 5,94E-02 & +\\
$F_{9}$ & 2,04E-08 & 2,04E-08 & 1,03E+02 & 4,07E+01 & + & 7,09E+01 & 2,48E+01 & + & 1,08E+02 & 3,12E+01 & +\\
$F_{10}$ & 1,46E-03 & 1,46E-03 & 3,41E+02 & 1,61E+01 & + & 3,17E+02 & 4,63E+01 & + & 3,47E+02 & 4,32E-01 & +\\ \hline
\end{tabular}
}
\end{table}

\begin{table}[!]
\centering
\caption{Statistical results of AHO in comparison to CSsin, MP-EEH, and RASPSHADE ($d=10$).}
\label{table12}
\resizebox{\textwidth}{!}{
\begin{tabular}{llllllllllll}
\hline
\multirow{2}{*}{} &
  \multicolumn{2}{c}{\textbf{AHO}} &
  \multicolumn{3}{c}{\textbf{CSsin}} &
  \multicolumn{3}{c}{\textbf{MP-EEH}} &
  \multicolumn{3}{c}{\textbf{RASPSHADE}} \\ \cline{2-12}
 &
  \multicolumn{1}{c}{\textbf{Mean}} &
  \multicolumn{1}{c}{\textbf{STD}} &
  \multicolumn{1}{c}{\textbf{Mean}} &
  \multicolumn{1}{c}{\textbf{STD}} &
  \multicolumn{1}{c}{\textbf{P}} &
  \multicolumn{1}{c}{\textbf{Mean}} &
  \multicolumn{1}{c}{\textbf{STD}} &
  \multicolumn{1}{c}{\textbf{P}} &
  \multicolumn{1}{c}{\textbf{Mean}} &
  \multicolumn{1}{c}{\textbf{STD}} &
  \multicolumn{1}{c}{\textbf{P}} \\ \hline
$F_{1}$ & 1,36E-08 & 1,36E-08 & 0,00E+00 & 0,00E+00 & - & 0,00E+00 & 0,00E+00 & - & 0,00E+00 & 0,00E+00 & -\\
$F_{2}$ & 1,27E-04 & 1,27E-04 & 1,58E+01 & 1,21E+01 & + & 3,48E+01 & 4,61E+01 & + & 9,44E-01 & 1,32E+00 & +\\
$F_{3}$ & 0,00E+00 & 0,00E+00 & 1,39E+01 & 1,71E+00 & + & 1,06E+01 & 3,62E+00 & + & 1,11E+01 & 3,58E-01 & +\\
$F_{4}$ & 0,00E+00 & 0,00E+00 & 0,00E+00 & 0,00E+00 &=& 8,67E-02 & 7,24E-02 & + & 2,72E-01 & 5,73E-02 & +\\
$F_{5}$ & 3,82E-05 & 3,82E-05 & 4,43E+00 & 1,82E+00 & + & 3,07E+01 & 2,09E+01 & + & 3,40E-01 & 1,73E-01 & +\\
$F_{6}$ & 4,58E-05 & 4,58E-05 & 1,70E-01 & 1,20E-01 & + & 1,23E+00 & 2,08E+00 & + & 1,59E-01 & 9,39E-02 & +\\
$F_{7}$ & 3,83E-05 & 3,83E-05 & 1,55E-01 & 9,95E-02 & + & 1,57E+00 & 1,86E+00 & + & 8,94E-04 & 9,58E-04 & +\\
$F_{8}$ & 1,09E-04 & 1,09E-04 & 8,01E+01 & 3,18E+01 & + & 4,67E+01 & 5,07E+01 & + & 1,00E+02 & 0,00E+00 & -\\
$F_{9}$ & 0,00E+00 & 0,00E+00 & 1,00E+02 & 1,38E-13 & + & 1,03E+02 & 1,83E+01 & + & 1,00E+02 & 0,00E+00 & =\\
$F_{10}$ & 3,37E-03 & 3,37E-03 & 3,77E+02 & 7,55E+01 & + & 3,38E+02 & 1,06E+02 & + & 3,98E+02 & 6,64E-02 & +\\ \hline
\end{tabular}
}
\end{table}

\begin{table}[!]
\centering
\caption{Statistical results of AHO in comparison to IMODE, DISH-XX, and AGSK ($d=10$).}
\label{table13}
\resizebox{\textwidth}{!}{
\begin{tabular}{llllllllllll}
\hline
\multirow{2}{*}{} &
  \multicolumn{2}{c}{\textbf{AHO}} &
  \multicolumn{3}{c}{\textbf{IMODE}} &
  \multicolumn{3}{c}{\textbf{DISH-XX}} &
  \multicolumn{3}{c}{\textbf{AGSK}} \\ \cline{2-12}
 &
  \multicolumn{1}{c}{\textbf{Mean}} &
  \multicolumn{1}{c}{\textbf{STD}} &
  \multicolumn{1}{c}{\textbf{Mean}} &
  \multicolumn{1}{c}{\textbf{STD}} &
  \multicolumn{1}{c}{\textbf{P}} &
  \multicolumn{1}{c}{\textbf{Mean}} &
  \multicolumn{1}{c}{\textbf{STD}} &
  \multicolumn{1}{c}{\textbf{P}} &
  \multicolumn{1}{c}{\textbf{Mean}} &
  \multicolumn{1}{c}{\textbf{STD}} &
  \multicolumn{1}{c}{\textbf{P}} \\ \hline
$F_{1}$ & 1,36E-08 & 1,36E-08 & 0,00E+00 & 0,00E+00 & - & 0,00E+00 & 0,00E+00 & - & 0,00E+00 & 0,00E+00 & -\\
$F_{2}$ & 1,27E-04 & 1,27E-04 & 4,20E+00 & 3,70E+00 & + & 1,97E+01 & 2,59E+01 & + & 2,84E+01 & 3,21E+01 & +\\
$F_{3}$ & 0,00E+00 & 0,00E+00 & 1,21E+01 & 7,83E-01 & + & 1,14E+01 & 5,85E-01 & + & 9,93E+00 & 4,26E+00 & +\\
$F_{4}$ & 0,00E+00 & 0,00E+00 & 0,00E+00 & 0,00E+00 &=& 2,47E-04 & 1,35E-03 & + & 5,83E-02 & 3,11E-02 & +\\
$F_{5}$ & 3,82E-05 & 3,82E-05 & 3,88E-01 & 3,83E-01 & + & 1,24E+00 & 2,80E+00 & + & 3,18E-01 & 3,06E-01 & +\\
$F_{6}$ & 4,58E-05 & 4,58E-05 & 9,15E-02 & 5,08E-02 & + & 6,96E-01 & 1,99E+00 & + & 1,55E-01 & 1,17E-01 & +\\
$F_{7}$ & 3,83E-05 & 3,83E-05 & 8,54E-04 & 1,10E-03 & + & 2,40E-01 & 2,50E-01 & + & 1,54E-03 & 1,71E-03 & +\\
$F_{8}$ & 1,09E-04 & 1,09E-04 & 2,72E+00 & 7,46E+00 & + & 1,00E+02 & 0,00E+00 & - & 1,80E+01 & 2,38E+01 & +\\
$F_{9}$ & 0,00E+00 & 0,00E+00 & 4,11E+01 & 4,46E+01 & + & 3,07E+02 & 7,01E+01 & + & 7,63E+01 & 4,29E+01 & +\\
$F_{10}$ & 3,37E-03 & 3,37E-03 & 3,98E+02 & 2,89E-13 & - & 4,07E+02 & 1,88E+01 & + & 2,98E+02 & 1,43E+02 & +\\ \hline
\end{tabular}
}
\end{table}

\begin{table}[!]
\centering
\caption{Statistical results of AHO in comparison to j2020, jDE100e, and OLSHADE ($d=10$).}
\label{table14}
\resizebox{\textwidth}{!}{
\begin{tabular}{llllllllllll}
\hline
\multirow{2}{*}{} &
  \multicolumn{2}{c}{\textbf{AHO}} &
  \multicolumn{3}{c}{\textbf{j2020}} &
  \multicolumn{3}{c}{\textbf{jDE100e}} &
  \multicolumn{3}{c}{\textbf{OLSHADE}} \\ \cline{2-12}
 &
  \multicolumn{1}{c}{\textbf{Mean}} &
  \multicolumn{1}{c}{\textbf{STD}} &
  \multicolumn{1}{c}{\textbf{Mean}} &
  \multicolumn{1}{c}{\textbf{STD}} &
  \multicolumn{1}{c}{\textbf{P}} &
  \multicolumn{1}{c}{\textbf{Mean}} &
  \multicolumn{1}{c}{\textbf{STD}} &
  \multicolumn{1}{c}{\textbf{P}} &
  \multicolumn{1}{c}{\textbf{Mean}} &
  \multicolumn{1}{c}{\textbf{STD}} &
  \multicolumn{1}{c}{\textbf{P}} \\ \hline
$F_{1}$ & 1,36E-08 & 1,36E-08 & 0,00E+00 & 0,00E+00 & - & 0,00E+00 & 0,00E+00 & - & 0,00E+00 & 0,00E+00 & -\\
$F_{2}$ & 1,27E-04 & 1,27E-04 & 6,79E-01 & 1,16E+00 & + & 1,01E+01 & 1,02E+01 & + & 2,92E+01 & 1,04E+01 & +\\
$F_{3}$ & 0,00E+00 & 0,00E+00 & 8,06E+00 & 3,88E+00 & + & 1,18E+01 & 1,83E+00 & + & 8,42E+00 & 3,23E+00 & +\\
$F_{4}$ & 0,00E+00 & 0,00E+00 & 1,09E-01 & 9,04E-02 & + & 1,65E-01 & 4,18E-02 & + & 8,74E-02 & 7,70E-02 & +\\
$F_{5}$ & 3,82E-05 & 3,82E-05 & 3,02E-01 & 3,13E-01 & + & 8,15E-01 & 5,75E-01 & + & 1,08E+00 & 1,91E+00 & +\\
$F_{6}$ & 4,58E-05 & 4,58E-05 & 4,78E-01 & 2,49E-01 & + & 3,46E-01 & 2,22E-01 & + & 0,00E+00 & 0,00E+00 & -\\
$F_{7}$ & 3,83E-05 & 3,83E-05 & 6,73E-02 & 1,25E-01 & + & 6,50E-03 & 7,87E-03 & + & 2,29E-01 & 2,63E-01 & +\\
$F_{8}$ & 1,09E-04 & 1,09E-04 & 1,54E+00 & 4,00E+00 & + & 3,33E+01 & 4,79E+01 & + & 3,36E+01 & 2,06E+01 & +\\
$F_{9}$ & 0,00E+00 & 0,00E+00 & 8,00E+01 & 4,07E+01 & + & 1,08E+02 & 4,19E+01 & + & 1,00E+02 & 0,00E+00 & =\\
$F_{10}$ & 3,37E-03 & 3,37E-03 & 1,40E+02 & 8,12E+01 & + & 3,98E+02 & 6,75E-02 & + & 1,00E+02 & 0,00E+00 & -\\ \hline
\end{tabular}
}
\end{table}

\begin{table}[!]
\centering
\caption{Statistical results of AHO in comparison to mpmLSHADE, SOMA-CL, and GSK ($d=10$).}
\label{table15}
\resizebox{\textwidth}{!}{
\begin{tabular}{llllllllllll}
\hline
\multirow{2}{*}{} &
  \multicolumn{2}{c}{\textbf{AHO}} &
  \multicolumn{3}{c}{\textbf{mpmLSHADE}} &
  \multicolumn{3}{c}{\textbf{SOMA-CL}} &
  \multicolumn{3}{c}{\textbf{GSK}} \\ \cline{2-12}
 &
  \multicolumn{1}{c}{\textbf{Mean}} &
  \multicolumn{1}{c}{\textbf{STD}} &
  \multicolumn{1}{c}{\textbf{Mean}} &
  \multicolumn{1}{c}{\textbf{STD}} &
  \multicolumn{1}{c}{\textbf{P}} &
  \multicolumn{1}{c}{\textbf{Mean}} &
  \multicolumn{1}{c}{\textbf{STD}} &
  \multicolumn{1}{c}{\textbf{P}} &
  \multicolumn{1}{c}{\textbf{Mean}} &
  \multicolumn{1}{c}{\textbf{STD}} &
  \multicolumn{1}{c}{\textbf{P}} \\ \hline
$F_{1}$ & 1,36E-08 & 1,36E-08 & 0,00E+00 & 0,00E+00 & - & 0,00E+00 & 0,00E+00 & - & 0,00E+00 & 0,00E+00 & -\\
$F_{2}$ & 1,27E-04 & 1,27E-04 & 6,81E-01 & 1,13E+00 & + & 1,47E+00 & 2,22E+00 & + & 8,19E+02 & 1,21E+02 & +\\
$F_{3}$ & 0,00E+00 & 0,00E+00 & 1,06E+01 & 2,34E-01 & + & 1,06E+01 & 2,27E-01 & + & 2,38E+01 & 2,84E+00 & +\\
$F_{4}$ & 0,00E+00 & 0,00E+00 & 2,84E-01 & 5,83E-02 & + & 1,52E-01 & 1,26E-01 & + & 1,46E+00 & 1,96E-01 & +\\
$F_{5}$ & 3,82E-05 & 3,82E-05 & 1,25E-01 & 1,15E-01 & + & 5,08E-01 & 2,02E+00 & + & 2,96E+01 & 7,44E+00 & +\\
$F_{6}$ & 4,58E-05 & 4,58E-05 & 5,09E-02 & 4,09E-02 & + & 1,72E-01 & 1,49E-01 & + & 2,69E+00 & 4,73E-01 & +\\
$F_{7}$ & 3,83E-05 & 3,83E-05 & 1,37E-01 & 1,55E-01 & + & 6,33E-02 & 1,51E-01 & + & 7,71E-01 & 1,71E-01 & +\\
$F_{8}$ & 1,09E-04 & 1,09E-04 & 9,67E+01 & 1,80E+01 & + & 5,71E+01 & 3,56E+01 & + & 9,71E+01 & 1,47E+01 & +\\
$F_{9}$ & 0,00E+00 & 0,00E+00 & 2,76E+02 & 9,82E+01 & + & 9,76E+01 & 1,91E+01 & + & 2,99E+02 & 4,99E+01 & +\\
$F_{10}$ & 3,37E-03 & 3,37E-03 & 4,03E+02 & 1,39E+01 & + & 4,00E+02 & 8,19E+00 & + & 3,99E+02 & 5,27E+00 & +\\ \hline
\end{tabular}
}
\end{table}

\begin{table}[!]
\centering
\caption{Statistical results of AHO in comparison to CSsin, MP-EEH, and RASPSHADE ($d=15$).}
\label{table16}
\resizebox{\textwidth}{!}{
\begin{tabular}{llllllllllll}
\hline
\multirow{2}{*}{} &
  \multicolumn{2}{c}{\textbf{AHO}} &
  \multicolumn{3}{c}{\textbf{CSsin}} &
  \multicolumn{3}{c}{\textbf{MP-EEH}} &
  \multicolumn{3}{c}{\textbf{RASPSHADE}} \\ \cline{2-12}
 &
  \multicolumn{1}{c}{\textbf{Mean}} &
  \multicolumn{1}{c}{\textbf{STD}} &
  \multicolumn{1}{c}{\textbf{Mean}} &
  \multicolumn{1}{c}{\textbf{STD}} &
  \multicolumn{1}{c}{\textbf{P}} &
  \multicolumn{1}{c}{\textbf{Mean}} &
  \multicolumn{1}{c}{\textbf{STD}} &
  \multicolumn{1}{c}{\textbf{P}} &
  \multicolumn{1}{c}{\textbf{Mean}} &
  \multicolumn{1}{c}{\textbf{STD}} &
  \multicolumn{1}{c}{\textbf{P}} \\ \hline
$F_{1}$ & 8,59E-08 & 8,59E-08 & 3,33E+08 & 1,82E+09 & + & 0,00E+00 & 0,00E+00 & - & 0,00E+00 & 0,00E+00 & -\\
$F_{2}$ & 1,91E-04 & 1,91E-04 & 7,25E+01 & 5,99E+01 & + & 1,15E+02 & 8,46E+01 & + & 1,02E+00 & 1,33E+00 & +\\
$F_{3}$ & 0,00E+00 & 0,00E+00 & 1,80E+01 & 1,25E+00 & + & 1,71E+01 & 5,03E+00 & + & 1,58E+01 & 2,29E-01 & +\\
$F_{4}$ & 0,00E+00 & 0,00E+00 & 0,00E+00 & 0,00E+00 &=& 2,89E-01 & 1,31E-01 & + & 3,64E-01 & 4,57E-02 & +\\
$F_{5}$ & 6,36E-05 & 6,36E-05 & 1,30E+01 & 6,19E+00 & + & 9,16E+01 & 4,80E+01 & + & 1,33E+00 & 8,38E-01 & +\\
$F_{6}$ & 5,64E-05 & 5,64E-05 & 1,44E+00 & 2,85E+00 & + & 4,07E+00 & 5,90E+00 & + & 6,17E-01 & 2,08E-01 & +\\
$F_{7}$ & 5,49E-05 & 5,49E-05 & 8,85E-01 & 1,96E-01 & + & 2,15E+01 & 3,89E+01 & + & 7,29E-01 & 1,24E-01 & +\\
$F_{8}$ & 1,52E-04 & 1,52E-04 & 8,75E+01 & 2,76E+01 & + & 6,00E+01 & 4,98E+01 & + & 1,00E+02 & 0,00E+00 & -\\
$F_{9}$ & 0,00E+00 & 0,00E+00 & 1,00E+02 & 2,30E-13 & + & 1,00E+02 & 2,19E-08 & + & 3,33E+02 & 4,08E+01 & +\\
$F_{10}$ & 9,07E-03 & 9,07E-03 & 4,00E+02 & 0,00E+00 & - & 3,90E+02 & 4,03E+01 & + & 4,00E+02 & 0,00E+00 & -\\ \hline
\end{tabular}
}
\end{table}

\begin{table}[!]
\centering
\caption{Statistical results of AHO in comparison to IMODE, DISH-XX, and AGSK ($d=15$).}
\label{table17}
\resizebox{\textwidth}{!}{
\begin{tabular}{llllllllllll}
\hline
\multirow{2}{*}{} &
  \multicolumn{2}{c}{\textbf{AHO}} &
  \multicolumn{3}{c}{\textbf{IMODE}} &
  \multicolumn{3}{c}{\textbf{DISH-XX}} &
  \multicolumn{3}{c}{\textbf{AGSK}} \\ \cline{2-12}
 &
  \multicolumn{1}{c}{\textbf{Mean}} &
  \multicolumn{1}{c}{\textbf{STD}} &
  \multicolumn{1}{c}{\textbf{Mean}} &
  \multicolumn{1}{c}{\textbf{STD}} &
  \multicolumn{1}{c}{\textbf{P}} &
  \multicolumn{1}{c}{\textbf{Mean}} &
  \multicolumn{1}{c}{\textbf{STD}} &
  \multicolumn{1}{c}{\textbf{P}} &
  \multicolumn{1}{c}{\textbf{Mean}} &
  \multicolumn{1}{c}{\textbf{STD}} &
  \multicolumn{1}{c}{\textbf{P}} \\ \hline
$F_{1}$ & 8,59E-08 & 8,59E-08 & 0,00E+00 & 0,00E+00 & - & 0,00E+00 & 0,00E+00 & - & 0,00E+00 & 0,00E+00 & -\\
$F_{2}$ & 1,91E-04 & 1,91E-04 & 3,14E+00 & 3,22E+00 & + & 7,13E+01 & 6,52E+01 & + & 1,85E+01 & 1,46E+01 & +\\
$F_{3}$ & 0,00E+00 & 0,00E+00 & 1,61E+01 & 3,12E-01 & + & 1,66E+01 & 5,43E-01 & + & 1,42E+01 & 4,27E+00 & +\\
$F_{4}$ & 0,00E+00 & 0,00E+00 & 0,00E+00 & 0,00E+00 &=& 0,00E+00 & 0,00E+00 &=& 1,42E-01 & 5,71E-02 & +\\
$F_{5}$ & 6,36E-05 & 6,36E-05 & 7,79E+00 & 3,66E+00 & + & 2,18E+01 & 3,31E+01 & + & 6,25E+00 & 4,32E+00 & +\\
$F_{6}$ & 5,64E-05 & 5,64E-05 & 6,92E-01 & 2,52E-01 & + & 9,40E+00 & 8,66E+00 & + & 4,02E-01 & 2,23E-01 & +\\
$F_{7}$ & 5,49E-05 & 5,49E-05 & 5,30E-01 & 2,23E-01 & + & 6,33E-01 & 2,51E-01 & + & 2,47E-01 & 2,00E-01 & +\\
$F_{8}$ & 1,52E-04 & 1,52E-04 & 4,18E+00 & 9,61E+00 & + & 1,00E+02 & 1,57E-13 & - & 6,85E+01 & 3,85E+01 & +\\
$F_{9}$ & 0,00E+00 & 0,00E+00 & 9,33E+01 & 2,54E+01 & + & 3,90E+02 & 8,27E-01 & + & 9,67E+01 & 1,83E+01 & +\\
$F_{10}$ & 9,07E-03 & 9,07E-03 & 4,00E+02 & 0,00E+00 & - & 4,00E+02 & 0,00E+00 & - & 4,00E+02 & 2,60E-13 & -\\ \hline
\end{tabular}
}
\end{table}

\begin{table}[!]
\centering
\caption{Statistical results of AHO in comparison to j2020, jDE100e, and OLSHADE ($d=15$).}
\label{table18}
\resizebox{\textwidth}{!}{
\begin{tabular}{llllllllllll}
\hline
\multirow{2}{*}{} &
  \multicolumn{2}{c}{\textbf{AHO}} &
  \multicolumn{3}{c}{\textbf{j2020}} &
  \multicolumn{3}{c}{\textbf{jDE100e}} &
  \multicolumn{3}{c}{\textbf{OLSHADE}} \\ \cline{2-12}
 &
  \multicolumn{1}{c}{\textbf{Mean}} &
  \multicolumn{1}{c}{\textbf{STD}} &
  \multicolumn{1}{c}{\textbf{Mean}} &
  \multicolumn{1}{c}{\textbf{STD}} &
  \multicolumn{1}{c}{\textbf{P}} &
  \multicolumn{1}{c}{\textbf{Mean}} &
  \multicolumn{1}{c}{\textbf{STD}} &
  \multicolumn{1}{c}{\textbf{P}} &
  \multicolumn{1}{c}{\textbf{Mean}} &
  \multicolumn{1}{c}{\textbf{STD}} &
  \multicolumn{1}{c}{\textbf{P}} \\ \hline
$F_{1}$ & 8,59E-08 & 8,59E-08 & 0,00E+00 & 0,00E+00 & - & 0,00E+00 & 0,00E+00 & - & 0,00E+00 & 0,00E+00 & -\\
$F_{2}$ & 1,91E-04 & 1,91E-04 & 5,72E-02 & 4,32E-02 & + & 2,60E+00 & 2,52E+00 & + & 1,10E+02 & 9,21E+01 & +\\
$F_{3}$ & 0,00E+00 & 0,00E+00 & 6,78E+00 & 7,82E+00 & + & 1,60E+01 & 4,02E-01 & + & 1,23E+01 & 6,03E+00 & +\\
$F_{4}$ & 0,00E+00 & 0,00E+00 & 1,99E-01 & 7,47E-02 & + & 2,57E-01 & 6,92E-02 & + & 5,37E-01 & 1,42E-01 & +\\
$F_{5}$ & 6,36E-05 & 6,36E-05 & 7,58E+00 & 7,69E+00 & + & 3,88E+00 & 2,29E+00 & + & 1,56E+01 & 4,06E+01 & +\\
$F_{6}$ & 5,64E-05 & 5,64E-05 & 8,45E-01 & 2,09E+00 & + & 4,79E-01 & 2,12E-01 & + & 0,00E+00 & 0,00E+00 & -\\
$F_{7}$ & 5,49E-05 & 5,49E-05 & 9,83E-01 & 2,03E+00 & + & 2,65E-01 & 1,38E-01 & + & 6,72E-01 & 1,82E-01 & +\\
$F_{8}$ & 1,52E-04 & 1,52E-04 & 9,49E+00 & 2,74E+01 & + & 1,00E+02 & 0,00E+00 & - & 1,00E+02 & 0,00E+00 & -\\
$F_{9}$ & 0,00E+00 & 0,00E+00 & 1,23E+02 & 5,68E+01 & + & 3,12E+02 & 1,30E+02 & + & 1,00E+02 & 0,00E+00 & =\\
$F_{10}$ & 9,07E-03 & 9,07E-03 & 3,90E+02 & 5,48E+01 & + & 4,00E+02 & 0,00E+00 & - & 1,70E+02 & 4,66E+01 & +\\ \hline
\end{tabular}
}
\end{table}

\begin{table}[!]
\centering
\caption{Statistical results of AHO in comparison to mpmLSHADE, SOMA-CL, and GSK ($d=15$).}
\label{table19}
\resizebox{\textwidth}{!}{
\begin{tabular}{llllllllllll}
\hline
\multirow{2}{*}{} &
  \multicolumn{2}{c}{\textbf{AHO}} &
  \multicolumn{3}{c}{\textbf{mpmLSHADE}} &
  \multicolumn{3}{c}{\textbf{SOMA-CL}} &
  \multicolumn{3}{c}{\textbf{GSK}} \\ \cline{2-12}
 &
  \multicolumn{1}{c}{\textbf{Mean}} &
  \multicolumn{1}{c}{\textbf{STD}} &
  \multicolumn{1}{c}{\textbf{Mean}} &
  \multicolumn{1}{c}{\textbf{STD}} &
  \multicolumn{1}{c}{\textbf{P}} &
  \multicolumn{1}{c}{\textbf{Mean}} &
  \multicolumn{1}{c}{\textbf{STD}} &
  \multicolumn{1}{c}{\textbf{P}} &
  \multicolumn{1}{c}{\textbf{Mean}} &
  \multicolumn{1}{c}{\textbf{STD}} &
  \multicolumn{1}{c}{\textbf{P}} \\ \hline
$F_{1}$ & 8,59E-08 & 8,59E-08 & 0,00E+00 & 0,00E+00 & - & 0,00E+00 & 0,00E+00 & - & 0,00E+00 & 0,00E+00 & -\\
$F_{2}$ & 1,91E-04 & 1,91E-04 & 1,83E-01 & 4,07E-01 & + & 1,82E+01 & 4,14E+01 & + & 1,78E+03 & 1,56E+02 & +\\
$F_{3}$ & 0,00E+00 & 0,00E+00 & 1,56E+01 & 2,10E-13 & + & 1,61E+01 & 4,64E-01 & + & 4,79E+01 & 3,88E+00 & +\\
$F_{4}$ & 0,00E+00 & 0,00E+00 & 3,82E-01 & 4,43E-02 & + & 7,04E-01 & 1,30E-01 & + & 3,29E+00 & 3,15E-01 & +\\
$F_{5}$ & 6,36E-05 & 6,36E-05 & 5,40E-01 & 5,04E-01 & + & 7,38E+00 & 4,50E+00 & + & 1,01E+02 & 1,65E+01 & +\\
$F_{6}$ & 5,64E-05 & 5,64E-05 & 7,25E-01 & 1,25E-01 & + & 6,52E+00 & 7,14E+00 & + & 4,62E+01 & 1,26E+01 & +\\
$F_{7}$ & 5,49E-05 & 5,49E-05 & 5,43E-01 & 1,15E-01 & + & 2,69E+00 & 7,48E+00 & + & 9,51E+00 & 2,39E+00 & +\\
$F_{8}$ & 1,52E-04 & 1,52E-04 & 1,00E+02 & 4,91E-13 & - & 8,07E+01 & 3,70E+01 & + & 1,00E+02 & 0,00E+00 & -\\
$F_{9}$ & 0,00E+00 & 0,00E+00 & 3,90E+02 & 4,66E-01 & + & 2,30E+02 & 1,19E+02 & + & 4,13E+02 & 4,51E+00 & +\\
$F_{10}$ & 9,07E-03 & 9,07E-03 & 4,00E+02 & 4,98E-13 & - & 4,00E+02 & 0,00E+00 & - & 4,00E+02 & 0,00E+00 & -\\ \hline
\end{tabular}
}
\end{table}

\begin{table}[!]
\centering
\caption{Statistical results of AHO in comparison to CSsin, MP-EEH, and RASPSHADE ($d=20$).}
\label{table20}
\resizebox{\textwidth}{!}{
\begin{tabular}{llllllllllll}
\hline
\multirow{2}{*}{} &
  \multicolumn{2}{c}{\textbf{AHO}} &
  \multicolumn{3}{c}{\textbf{CSsin}} &
  \multicolumn{3}{c}{\textbf{MP-EEH}} &
  \multicolumn{3}{c}{\textbf{RASPSHADE}} \\ \cline{2-12}
 &
  \multicolumn{1}{c}{\textbf{Mean}} &
  \multicolumn{1}{c}{\textbf{STD}} &
  \multicolumn{1}{c}{\textbf{Mean}} &
  \multicolumn{1}{c}{\textbf{STD}} &
  \multicolumn{1}{c}{\textbf{P}} &
  \multicolumn{1}{c}{\textbf{Mean}} &
  \multicolumn{1}{c}{\textbf{STD}} &
  \multicolumn{1}{c}{\textbf{P}} &
  \multicolumn{1}{c}{\textbf{Mean}} &
  \multicolumn{1}{c}{\textbf{STD}} &
  \multicolumn{1}{c}{\textbf{P}} \\ \hline
$F_{1}$ & 5,09E-07 & 5,09E-07 & 9,33E+09 & 2,53E+09 & + & 0,00E+00 & 0,00E+00 & - & 0,00E+00 & 0,00E+00 & -\\
$F_{2}$ & 2,55E-04 & 2,55E-04 & 9,83E+01 & 8,33E+01 & + & 1,70E+02 & 9,42E+01 & + & 1,38E-01 & 4,54E-02 & +\\
$F_{3}$ & 0,00E+00 & 0,00E+00 & 2,55E+01 & 2,27E+00 & + & 2,33E+01 & 6,13E+00 & + & 2,05E+01 & 1,89E-01 & +\\
$F_{4}$ & 0,00E+00 & 0,00E+00 & 0,00E+00 & 0,00E+00 &=& 4,25E-01 & 1,41E-01 & + & 4,53E-01 & 4,18E-02 & +\\
$F_{5}$ & 7,67E-05 & 7,67E-05 & 1,16E+02 & 6,34E+01 & + & 2,36E+02 & 7,80E+01 & + & 1,41E+00 & 1,25E+00 & +\\
$F_{6}$ & 1,00E-04 & 1,00E-04 & 6,72E-01 & 8,22E+00 & + & 4,27E+01 & 5,23E+01 & + & 1,71E-01 & 5,74E-02 & +\\
$F_{7}$ & 6,74E-05 & 6,74E-05 & 2,62E+00 & 2,26E+00 & + & 7,77E+01 & 6,52E+01 & + & 7,27E-01 & 2,40E-01 & +\\
$F_{8}$ & 2,00E-04 & 2,00E-04 & 9,89E+01 & 5,59E+00 & + & 8,00E+01 & 4,07E+01 & + & 1,00E+02 & 0,00E+00 & -\\
$F_{9}$ & 0,00E+00 & 0,00E+00 & 1,03E+02 & 1,82E+01 & + & 9,67E+01 & 1,83E+01 & + & 3,42E+02 & 3,51E+01 & +\\
$F_{10}$ & 8,83E-03 & 8,83E-03 & 3,99E+02 & 2,44E-01 & + & 4,01E+02 & 3,78E+00 & + & 4,14E+02 & 1,14E-13 & -\\ \hline
\end{tabular}
}
\end{table}

\begin{table}[!]
\centering
\caption{Statistical results of AHO in comparison to IMODE, DISH-XX, and AGSK ($d=20$).}
\label{table21}
\resizebox{\textwidth}{!}{
\begin{tabular}{llllllllllll}
\hline
\multirow{2}{*}{} &
  \multicolumn{2}{c}{\textbf{AHO}} &
  \multicolumn{3}{c}{\textbf{IMODE}} &
  \multicolumn{3}{c}{\textbf{DISH-XX}} &
  \multicolumn{3}{c}{\textbf{AGSK}} \\ \cline{2-12}
 &
  \multicolumn{1}{c}{\textbf{Mean}} &
  \multicolumn{1}{c}{\textbf{STD}} &
  \multicolumn{1}{c}{\textbf{Mean}} &
  \multicolumn{1}{c}{\textbf{STD}} &
  \multicolumn{1}{c}{\textbf{P}} &
  \multicolumn{1}{c}{\textbf{Mean}} &
  \multicolumn{1}{c}{\textbf{STD}} &
  \multicolumn{1}{c}{\textbf{P}} &
  \multicolumn{1}{c}{\textbf{Mean}} &
  \multicolumn{1}{c}{\textbf{STD}} &
  \multicolumn{1}{c}{\textbf{P}} \\ \hline
$F_{1}$ & 5,09E-07 & 5,09E-07 & 0,00E+00 & 0,00E+00 & - & 0,00E+00 & 0,00E+00 & - & 0,00E+00 & 0,00E+00 & -\\
$F_{2}$ & 2,55E-04 & 2,55E-04 & 5,13E-01 & 7,13E-01 & + & 8,67E+01 & 1,11E+02 & + & 9,68E-01 & 1,23E+00 & +\\
$F_{3}$ & 0,00E+00 & 0,00E+00 & 2,05E+01 & 1,26E-01 & + & 2,13E+01 & 3,74E+00 & + & 2,04E+01 & 0,00E+00 & =\\
$F_{4}$ & 0,00E+00 & 0,00E+00 & 0,00E+00 & 0,00E+00 &=& 2,47E-04 & 1,35E-03 & + & 1,45E-01 & 5,47E-02 & +\\
$F_{5}$ & 7,67E-05 & 7,67E-05 & 1,09E+01 & 4,33E+00 & + & 5,63E+01 & 6,63E+01 & + & 4,50E+01 & 3,67E+01 & +\\
$F_{6}$ & 1,00E-04 & 1,00E-04 & 3,02E-01 & 8,17E-02 & + & 1,50E+01 & 3,57E+01 & + & 1,68E-01 & 4,45E-02 & +\\
$F_{7}$ & 6,74E-05 & 6,74E-05 & 5,24E-01 & 1,64E-01 & + & 5,09E+00 & 6,42E+00 & + & 6,81E-01 & 9,09E-01 & +\\
$F_{8}$ & 2,00E-04 & 2,00E-04 & 8,40E+01 & 1,89E+01 & + & 1,00E+02 & 1,39E-13 & - & 9,92E+01 & 4,63E+00 & +\\
$F_{9}$ & 0,00E+00 & 0,00E+00 & 9,67E+01 & 1,83E+01 & + & 4,05E+02 & 2,50E+00 & + & 1,00E+02 & 8,30E-14 & +\\
$F_{10}$ & 8,83E-03 & 8,83E-03 & 4,00E+02 & 6,18E-01 & + & 4,14E+02 & 2,54E-02 & + & 3,99E+02 & 1,59E-02 & +\\ \hline
\end{tabular}
}
\end{table}

\begin{table}[!]
\centering
\caption{Statistical results of AHO in comparison to j2020, jDE100e, and OLSHADE ($d=20$).}
\label{table22}
\resizebox{\textwidth}{!}{
\begin{tabular}{llllllllllll}
\hline
\multirow{2}{*}{} &
  \multicolumn{2}{c}{\textbf{AHO}} &
  \multicolumn{3}{c}{\textbf{j2020}} &
  \multicolumn{3}{c}{\textbf{jDE100e}} &
  \multicolumn{3}{c}{\textbf{OLSHADE}} \\ \cline{2-12}
 &
  \multicolumn{1}{c}{\textbf{Mean}} &
  \multicolumn{1}{c}{\textbf{STD}} &
  \multicolumn{1}{c}{\textbf{Mean}} &
  \multicolumn{1}{c}{\textbf{STD}} &
  \multicolumn{1}{c}{\textbf{P}} &
  \multicolumn{1}{c}{\textbf{Mean}} &
  \multicolumn{1}{c}{\textbf{STD}} &
  \multicolumn{1}{c}{\textbf{P}} &
  \multicolumn{1}{c}{\textbf{Mean}} &
  \multicolumn{1}{c}{\textbf{STD}} &
  \multicolumn{1}{c}{\textbf{P}} \\ \hline
$F_{1}$ & 5,09E-07 & 5,09E-07 & 0,00E+00 & 0,00E+00 & - & 0,00E+00 & 0,00E+00 & - & 0,00E+00 & 0,00E+00 & -\\
$F_{2}$ & 2,55E-04 & 2,55E-04 & 2,60E-02 & 2,47E-02 & + & 1,48E+00 & 1,50E+00 & + & 1,15E+00 & 7,62E+01 & +\\
$F_{3}$ & 0,00E+00 & 0,00E+00 & 1,44E+01 & 9,29E+00 & + & 2,10E+01 & 4,09E-01 & + & 2,52E+00 & 7,63E+00 & +\\
$F_{4}$ & 0,00E+00 & 0,00E+00 & 1,80E-01 & 7,84E-02 & + & 3,47E-01 & 8,04E-02 & + & 1,01E+00 & 1,22E+00 & +\\
$F_{5}$ & 7,67E-05 & 7,67E-05 & 7,78E+01 & 5,75E+01 & + & 2,37E+00 & 8,49E-01 & + & 1,78E+00 & 4,14E+01 & +\\
$F_{6}$ & 1,00E-04 & 1,00E-04 & 1,92E-01 & 1,01E-01 & + & 1,15E-01 & 3,31E-02 & + & 5,17E-01 & 0,00E+00 & -\\
$F_{7}$ & 6,74E-05 & 6,74E-05 & 1,98E+00 & 4,02E+00 & + & 2,14E-01 & 1,14E-01 & + & 8,42E-01 & 1,61E-01 & +\\
$F_{8}$ & 2,00E-04 & 2,00E-04 & 9,27E+01 & 2,21E+01 & + & 1,00E+02 & 0,00E+00 & - & 1,00E+00 & 7,01E-01 & +\\
$F_{9}$ & 0,00E+00 & 0,00E+00 & 3,39E+02 & 1,28E+02 & + & 4,05E+02 & 1,74E+00 & + & 1,10E+00 & 3,06E+01 & +\\
$F_{10}$ & 8,83E-03 & 8,83E-03 & 3,99E+02 & 4,02E-02 & + & 4,13E+02 & 2,78E+00 & + & 4,10E+00 & 6,15E+00 & +\\ \hline
\end{tabular}
}
\end{table}

\begin{table}[!]
\centering
\caption{Statistical results of AHO in comparison to mpmLSHADE, SOMA-CL, and GSK ($d=20$).}
\label{table23}
\resizebox{\textwidth}{!}{
\begin{tabular}{llllllllllll}
\hline
\multirow{2}{*}{} &
  \multicolumn{2}{c}{\textbf{AHO}} &
  \multicolumn{3}{c}{\textbf{mpmLSHADE}} &
  \multicolumn{3}{c}{\textbf{SOMA-CL}} &
  \multicolumn{3}{c}{\textbf{GSK}} \\ \cline{2-12}
 &
  \multicolumn{1}{c}{\textbf{Mean}} &
  \multicolumn{1}{c}{\textbf{STD}} &
  \multicolumn{1}{c}{\textbf{Mean}} &
  \multicolumn{1}{c}{\textbf{STD}} &
  \multicolumn{1}{c}{\textbf{P}} &
  \multicolumn{1}{c}{\textbf{Mean}} &
  \multicolumn{1}{c}{\textbf{STD}} &
  \multicolumn{1}{c}{\textbf{P}} &
  \multicolumn{1}{c}{\textbf{Mean}} &
  \multicolumn{1}{c}{\textbf{STD}} &
  \multicolumn{1}{c}{\textbf{P}} \\ \hline
$F_{1}$ & 5,09E-07 & 5,09E-07 & 0,00E+00 & 0,00E+00 & - & 0,00E+00 & 0,00E+00 & - & 0,00E+00 & 0,00E+00 & -\\
$F_{2}$ & 2,55E-04 & 2,55E-04 & 3,97E-02 & 2,12E-02 & + & 7,36E+00 & 2,15E+01 & + & 2,68E+03 & 1,60E+02 & +\\
$F_{3}$ & 0,00E+00 & 0,00E+00 & 2,04E+01 & 4,67E-13 & + & 2,14E+01 & 8,06E-01 & + & 7,37E+01 & 5,25E+00 & +\\
$F_{4}$ & 0,00E+00 & 0,00E+00 & 4,97E-01 & 4,23E-02 & + & 1,05E+00 & 1,83E-01 & + & 5,37E+00 & 4,25E-01 & +\\
$F_{5}$ & 7,67E-05 & 7,67E-05 & 1,38E+00 & 1,45E+00 & + & 1,45E+02 & 8,01E+01 & + & 2,44E+02 & 3,97E+01 & +\\
$F_{6}$ & 1,00E-04 & 1,00E-04 & 2,05E-01 & 4,71E-02 & + & 2,71E-01 & 8,35E-02 & + & 3,35E+00 & 2,17E+00 & +\\
$F_{7}$ & 6,74E-05 & 6,74E-05 & 5,10E-01 & 1,19E-01 & + & 9,22E+00 & 8,93E+00 & + & 5,86E+01 & 1,09E+01 & +\\
$F_{8}$ & 2,00E-04 & 2,00E-04 & 1,00E+02 & 7,47E-13 & - & 9,91E+01 & 3,75E+00 & + & 1,00E+02 & 0,00E+00 & -\\
$F_{9}$ & 0,00E+00 & 0,00E+00 & 4,01E+02 & 6,68E-01 & + & 3,99E+02 & 4,54E+01 & + & 4,39E+02 & 2,95E+01 & +\\
$F_{10}$ & 8,83E-03 & 8,83E-03 & 4,14E+02 & 2,74E-04 & - & 4,71E+02 & 3,23E+01 & + & 4,14E+02 & 8,87E-03 & +\\ \hline
\end{tabular}
}
\end{table}

\begin{table}[!]
\centering
\caption{The Wilcoxon signed rank test for $d=5$}
\label{table24}
\resizebox{\textwidth}{!}{
\begin{tabular}{lllllll}
\hline
 & $k$ & $W+$ & $W-$ & $\min(W+,W-)$ & Critical value & Result of the Wilcoxon signed rank test\\ \hline
AHO vs. CSsin & 7 & 0 & 28 & 0 & 4 & $4>0$, AHO outperforms CSsin \\
AHO vs. MP-EEH & 7 & 1 & 27 & 1 & 4 & $4>1$, AHO outperforms MP-EEH \\
AHO vs. RASPSHADE & 7 & 3 & 25 & 3 & 4 & $4>3$, AHO outperforms RASPSHADE \\
AHO vs. IMODE & 5 & 6 & 9 & 6 & 1 & $1<6$, IMODE outperforms AHO \\
AHO vs. DISH-XX & 7 & 0 & 28 & 0 & 4 & $4>0$, AHO outperforms DISH-XX \\
AHO vs. AGSK & 7 & 3 & 25 & 3 & 4 & $4>3$, AHO outperforms AGSK \\
AHO vs. j2020 & 7 & 0 & 28 & 0 & 4 & $4>0$, AHO outperforms j2020 \\
AHO vs. jDE100e & 7 & 0 & 28 & 0 & 4 & $4>0$, AHO outperforms jDE100e \\
AHO vs. OLSHADE & 7 & 3 & 25 & 3 & 4 & $4>3$, AHO outperforms OLSHADE \\
AHO vs. mpmLSHADE & 7 & 1 & 27 & 1 & 4 & $4>1$, AHO outperforms mpmLSHADE \\
AHO vs. SOMA-CL & 8 & 0 & 36 & 0 & 6 & $6>0$, AHO outperforms SOMA-CL \\
AHO vs. GSK & 7 & 0 & 28 & 0 & 4 & $4>0$, AHO outperforms GSK \\ \hline
\end{tabular}
}
\end{table}

\begin{table}[!]
\centering
\caption{The Wilcoxon signed rank test for $d=10$}
\label{table25}
\resizebox{\textwidth}{!}{
\begin{tabular}{lllllll}
\hline
 & $k$ & $W+$ & $W-$ & $\min(W+,W-)$ & Critical value & Result of the Wilcoxon signed rank test\\ \hline
AHO vs. CSsin & 9 & 2 & 43 & 2 & 8 & $8 > 2$, AHO outperforms CSsin \\
AHO vs. MP-EEH & 10 & 1 & 54 & 1 & 11 & $11 > 1$, AHO outperforms MP-EEH \\
AHO vs. RASPSHADE & 9 & 3 & 42 & 3 & 8 & $8 > 3$, AHO outperforms RASPSHADE \\
AHO vs. IMODE & 9 & 4 & 41 & 4 & 8 & $8 > 4$, AHO outperforms IMODE \\
AHO vs. DISH-XX & 10 & 3 & 52 & 3 & 11 & $11 > 3$, AHO outperforms DISH-XX \\
AHO vs. AGSK & 10 & 1 & 54 & 1 & 11 & $11 > 1$, AHO outperforms AGSK\\
AHO vs. j2020 & 10 & 1 & 54 & 1 & 11 & $11 > 1$, AHO outperforms j2020\\
AHO vs. jDE100e & 10 & 1 & 54 & 1 & 11 & $11 > 1$, AHO outperforms jDE100e\\
AHO vs. OLSHADE & 9 & 6 & 39 & 6 & 8 & $8 > 6$, AHO outperforms OLSHADE\\
AHO vs. mpmLSHADE & 10 & 1 & 54 & 1 & 11 & $11 > 1$, AHO outperforms mpmLSHADE\\
AHO vs. SOMA-CL & 10 & 1 & 54 & 1 & 11 & $11 > 1$, AHO outperforms SOMA-CL\\
AHO vs. GSK & 10 & 1 & 54 & 1 & 11 & $11 > 1$, AHO outperforms GSK \\ \hline
\end{tabular}
}
\end{table}

\begin{table}[!]
\centering
\caption{The Wilcoxon signed rank test for $d=15$}
\label{table26}
\resizebox{\textwidth}{!}{
\begin{tabular}{lllllll}
\hline
 & $k$ & $W+$ & $W-$ & $\min(W+,W-)$ & Critical value & Result of the Wilcoxon signed rank test\\ \hline
AHO vs. CSsin & 9 & 2 & 43 & 2 & 8 & $8 > 2$, AHO outperforms CSsin \\
AHO vs. MP-EEH & 10 & 2 & 53 & 2 & 11 & $11 > 2$, AHO outperforms MP-EEH \\
AHO vs. RASPSHADE & 10 & 6 & 49 & 6 & 11 & $11 > 6$, AHO outperforms RASPSHADE \\
AHO vs. IMODE & 9 & 3 & 42 & 3 & 8 & $8 > 3$, AHO outperforms IMODE \\
AHO vs. DISH-XX & 9 & 6 & 39 & 6 & 8 & $8 > 6$, AHO outperforms DISH-XX \\
AHO vs. AGSK & 10 & 3 & 52 & 3 & 11 & $11 > 3$, AHO outperforms AGSK \\
AHO vs. j2020 & 10 & 1 & 54 & 1 & 11 & $11 > 1$, AHO outperforms j2020 \\
AHO vs. jDE100e & 10 & 6 & 49 & 6 & 11 & $11 > 6$, AHO outperforms jDE100e \\
AHO vs. OLSHADE & 9 & 6 & 39 & 6 & 8 & $8 > 6$, AHO outperforms OLSHADE \\
AHO vs. mpmLSHADE & 10 & 9 & 46 & 9 & 11 & $11 > 9$, AHO outperforms mpmLSHADE \\
AHO vs. SOMA-CL & 10 & 3 & 52 & 3 & 11 & $11 > 3$, AHO outperforms SOMA-CL \\
AHO vs. GSK & 10 & 6 & 49 & 6 & 11 & $11 > 6$, AHO outperforms GSK \\ \hline
\end{tabular}
}
\end{table}

\begin{table}[!]
\centering
\caption{The Wilcoxon signed rank test for $d=20$}
\label{table27}
\resizebox{\textwidth}{!}{
\begin{tabular}{lllllll}
\hline
 & $k$ & $W+$ & $W-$ & $\min(W+,W-)$ & Critical value & Result of the Wilcoxon signed rank test\\ \hline
AHO vs. CSsin & 9 & 0 & 45 & 0 & 8 & $8 > 0$, AHO outperforms CSsin \\
AHO vs. MP-EEH & 10 & 1 & 54 & 1 & 11 & $11 > 1$, AHO outperforms MP-EEH \\
AHO vs. RASPSHADE & 10 & 6 & 49 & 6 & 11 & $11 > 6$, AHO outperforms RASPSHADE \\
AHO vs. IMODE & 9 & 1 & 44 & 1 & 8 & $8 > 1$, AHO outperforms IMODE \\
AHO vs. DISH-XX & 10 & 3 & 52 & 3 & 11 & $11 > 3$, AHO outperforms DISH-XX \\
AHO vs. AGSK & 9 & 2 & 43 & 2 & 8 & $8 > 2$, AHO outperforms AGSK \\
AHO vs. j2020 & 10 & 1 & 54 & 1 & 11 & $11 > 1$, AHO outperforms j2020 \\
AHO vs. jDE100e & 10 & 3 & 52 & 3 & 11 & $11 > 3$, AHO outperforms jDE100e \\
AHO vs. OLSHADE & 10 & 3 & 52 & 3 & 11 & $11 > 3$, AHO outperforms OLSHADE \\
AHO vs. mpmLSHADE & 10 & 9 & 46 & 9 & 11 & $11 > 9$, AHO outperforms mpmLSHADE \\
AHO vs. SOMA-CL & 10 & 1 & 54 & 1 & 11 & $11 > 1$, AHO outperforms SOMA-CL \\
AHO vs. GSK & 10 & 4 & 51 & 4 & 11 & $11 > 4$, AHO outperforms GSK \\ \hline
\end{tabular}
}
\end{table}

From Tables \ref{table24}, \ref{table25}, \ref{table26}, and \ref{table27}, AHO successfully outperforms all the metaheuristic algorithms considered for the comparative study in all dimension, except for the dimension $d=5$ where AHO is outperformed by IMODE.

\begin{figure}[!]
\centering
\begin{subfigure}[b]{0.1\textwidth}
\centering
\includegraphics[scale=0.3]{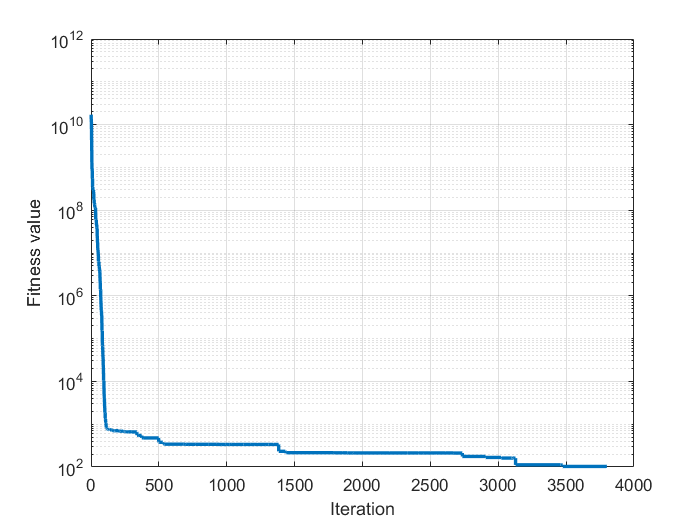}
\caption{$F_{1}$}
\end{subfigure}
\hfill
\begin{subfigure}[b]{0.5\textwidth}
\centering
\includegraphics[scale=0.3]{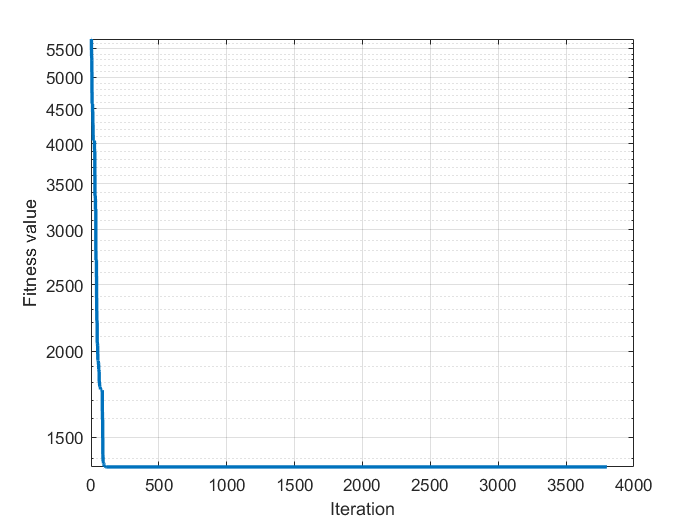}
\caption{$F_{2}$}
\end{subfigure}
\vfill
\begin{subfigure}[b]{0.1\textwidth}
\centering
\includegraphics[scale=0.3]{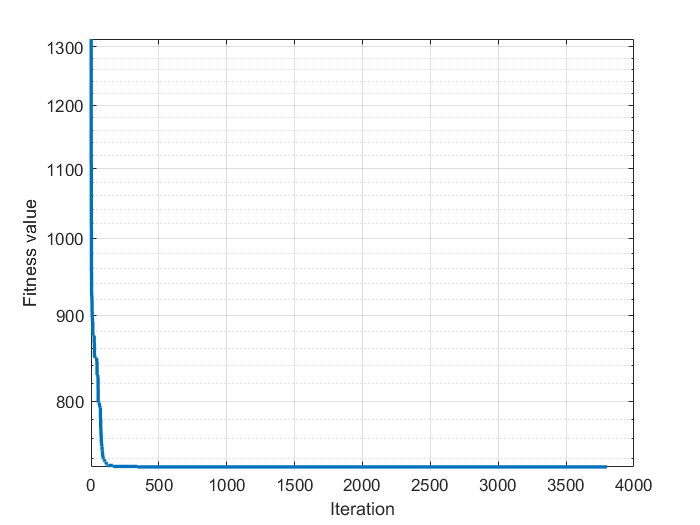}
\caption{$F_{3}$}
\end{subfigure}
\hfill
\begin{subfigure}[b]{0.5\textwidth}
\centering
\includegraphics[scale=0.3]{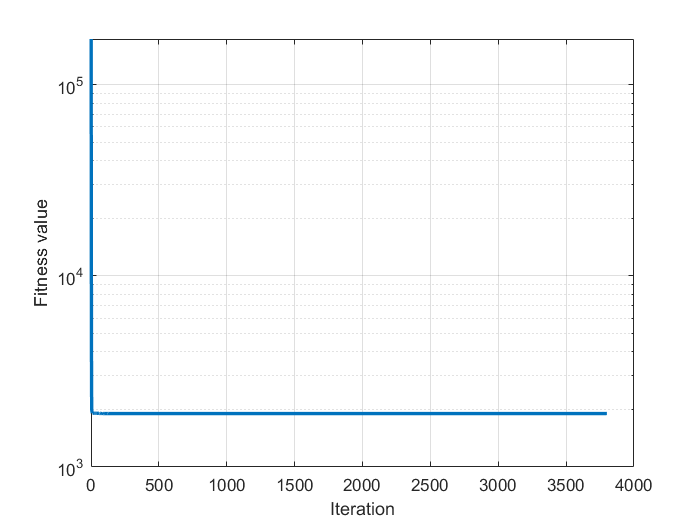}
\caption{$F_{4}$}
\end{subfigure}
\vfill\begin{subfigure}[b]{0.1\textwidth}
\centering
\includegraphics[scale=0.3]{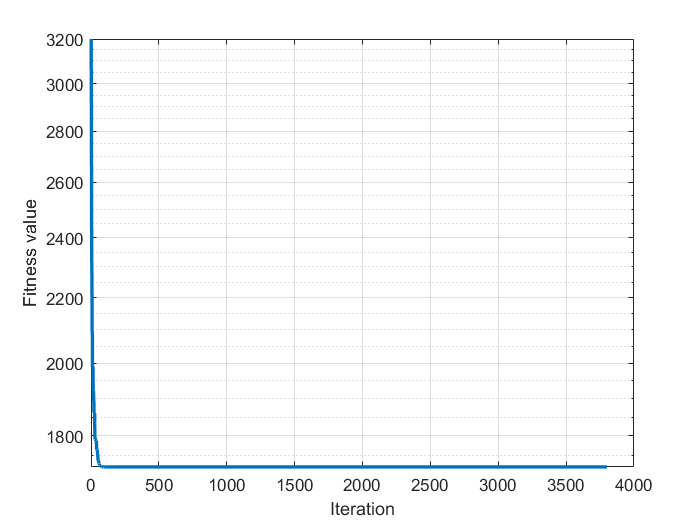}
\caption{$F_{5}$}
\end{subfigure}
\hfill
\begin{subfigure}[b]{0.5\textwidth}
\centering
\includegraphics[scale=0.3]{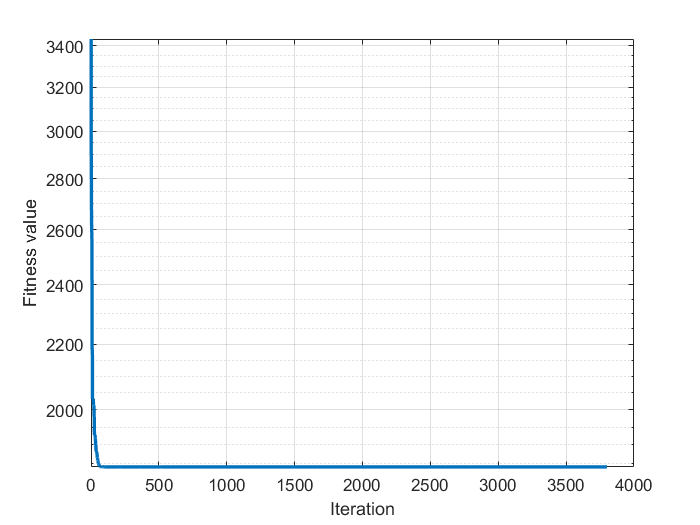}
\caption{$F_{6}$}
\end{subfigure}
\vfill\begin{subfigure}[b]{0.1\textwidth}
\centering
\includegraphics[scale=0.3]{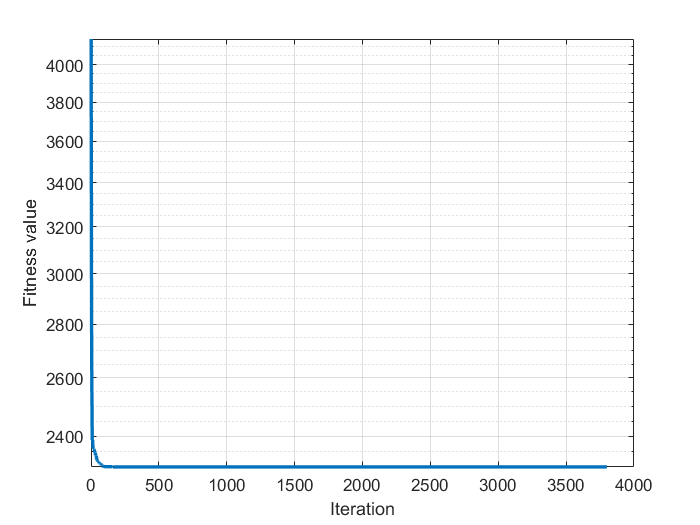}
\caption{$F_{7}$}
\end{subfigure}
\hfill
\begin{subfigure}[b]{0.5\textwidth}
\centering
\includegraphics[scale=0.3]{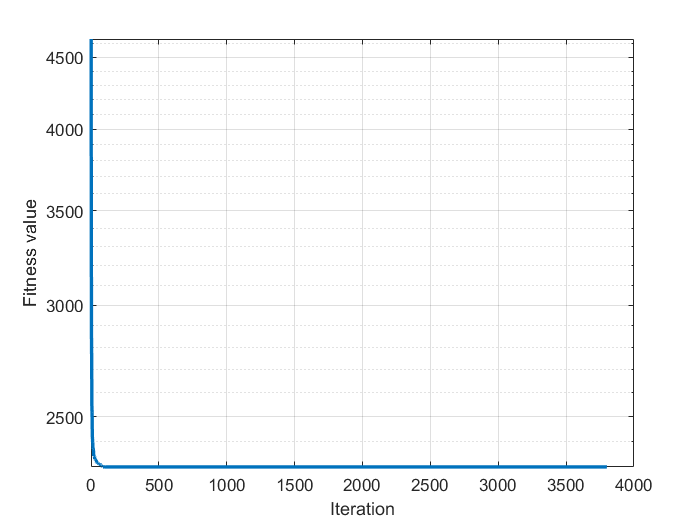}
\caption{$F_{8}$}
\end{subfigure}
\vfill\begin{subfigure}[b]{0.1\textwidth}
\centering
\includegraphics[scale=0.3]{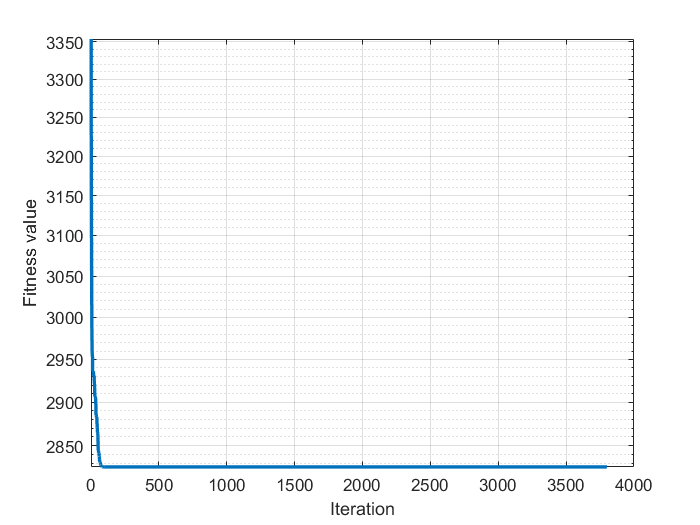}
\caption{$F_{9}$}
\end{subfigure}
\hfill
\begin{subfigure}[b]{0.5\textwidth}
\centering
\includegraphics[scale=0.3]{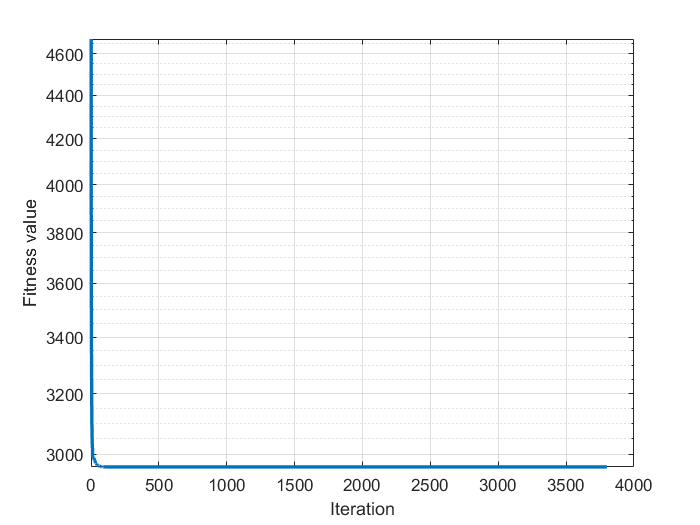}
\caption{$F_{10}$}
\end{subfigure}
\caption{Convergence curves of AHO for test functions of CEC 2020 in 20d, averaged over 30 runs.}
\label{figure5}
\end{figure}

Figure \ref{figure5} shows AHO's averaged convergence curves for the ten test functions of the benchmark CEC 2020 when $d=20$, over 30 independent runs. We observe that the exploitation with AHO on most test functions is dominant, leading to faster convergence to the global optimum because the exploitation phase is strengthened by Equations \ref{equation4} and \ref{equation5}.

\section{Experimental results on constrained optimization problems}
\label{section5}

We consider five constrained engineering design problems \cite{kumar2020test} to evaluate AHO's performance. Their characteristics are reported in Table \ref{table28}, while the complete mathematical equations of these problems are provided in the following subsections.

\begin{table}[!]
\centering
\caption{Characteristics of the considered constrained engineering design problems.}
\label{table28}
\begin{tabular}{llllll}
\hline
Problem & Name & $D$ & $g$ & $h$ & $f(x^*)$ \\ \hline
RC15 & Weight Minimization of a Speed Reducer & 7 & 11 & 0 & 2.9944244658E+03 \\
RC18 & Pressure vessel design & 4 & 4 & 0 & 5.8853327736E+03 \\
RC19 & Welded beam design & 4 & 5 & 0 & 1.6702177263E+00 \\
RC17 & Tension/compression spring design (case 1) & 3 & 3 & 0 & 1.2665232788E-02 \\
RC21 & Multiple disk clutch brake design problem & 5 & 7 & 0 & 2.3524245790E-01 \\ \hline
\multicolumn{6}{l}{$D$: the number of decision variables of the problem.} \\
\multicolumn{6}{l}{$g$: the number of inequality constraints.} \\
\multicolumn{6}{l}{$h$: the number of equality constraints.} \\
\multicolumn{6}{l}{$f(x^*)$: the best known feasible objective function value.} \\ \hline
\end{tabular}
\end{table}

All the experiments were run using the Java programming language on a workstation with a Windows 10 familial edition (64-bit). The processor is Intel(R) Core(TM) i7--9750H CPU @ 2.60GHz 2.59 GHz, with 16 GB of RAM. The dimension of the search space ($D$) of each problem is given in Table \ref{table28}. The population size ($N$) is set to $\lfloor 30 \times D^{1.5} \rfloor$, where the term $\lfloor x \rfloor$ returns the truncation of the real number $x$. For each problem, the maximum number of iterations ($Iter_{\max}$) is equal to $\frac{200000}{N}$. The range of allowed values for each decision variable is given in the following subsections. All the results are averaged over 25 independent runs. The swapping angle value ($\theta$) is set to $\frac{\pi}{12}$. The attractiveness rate value ($\omega$) is estimated to 0.01. For comparative purposes, the results of 3 well-established metaheuristic algorithms from the literature are presented to have a conclusive decision on AHO's performance. The following metaheuristic algorithms are used for the comparative study.

\begin{itemize}
  \item An improved unified differential evolution algorithm for constrained optimization problems (IUDE) \cite{trivedi2018improved}.
  \item A matrix adaptation evolution strategy for constrained real-parameter optimization ($\epsilon$MAgES) \cite{hellwig2018matrix}.
  \item LSHADE44 with an Improved $\epsilon$ Constraint-Handling Method for Solving Constrained Single-Objective Optimization Problems (iLSHADE$_{\epsilon}$) \cite{fan2018lshade44}.
\end{itemize}

\begin{table}[!]
\centering
\caption{Results of the considered mechanical engineering problems using IUDE, $\epsilon$MAgES, iLSHADE$_{\epsilon}$, and AHO.}
\label{table29}
\resizebox{\textwidth}{!}{
\begin{tabular}{llllllllll}
\hline
\multicolumn{1}{c}{\textbf{Problem}} &
  \multicolumn{1}{c}{\textbf{Algorithm}} &
  \multicolumn{1}{c}{\textbf{Best}} &
  \multicolumn{1}{c}{\textbf{Median}} &
  \multicolumn{1}{c}{\textbf{Mean}} &
  \multicolumn{1}{c}{\textbf{Worst}} &
  \multicolumn{1}{c}{\textbf{STD}} &
  \multicolumn{1}{c}{\textbf{FR}} &
  \multicolumn{1}{c}{\textbf{MV}} &
  \multicolumn{1}{c}{\textbf{SR}} \\ \hline
\multirow{4}{*}{\textbf{RC15}} & \textbf{IUDE} & 2.99E+03 & 2.99E+03 & 2.99E+03 & 2.99E+03 & 4.64E-13 & 100 & 0.00E+00 & 100 \\
                               & \textbf{$\epsilon$MAgES} & 2.99E+03 & 2.99E+03 & 2.99E+03 & 2.99E+03 & 4.64E-13 & 100 & 0.00E+00 & 100 \\
                               & \textbf{iLSHADE$_{\epsilon}$} & 2.99E+03 & 2.99E+03 & 2.99E+03 & 2.99E+03 & 4.64E-13 & 100 & 0.00E+00 & 100 \\
                               & \textbf{AHO} & 2.99E+03 & 2.99E+03 & 2.99E+03 & 2.99E+03 & 4.64E-13 & 100 & 0.00E+00 & 100 \\ \hline
\multirow{4}{*}{\textbf{RC18}} & \textbf{IUDE} &  6.06E+03 & 6.06E+03 & 6.06E+03 & 6.09E+03 & 6.16E+00 & 100 & 0.00E+00 & 24 \\
                               & \textbf{$\epsilon$MAgES} &  6.06E+03 & 6.41E+03 & 7.38E+03 & 1.19E+04 & 1.93E+03 & 100 & 0.00E+00 & 16 \\
                               & \textbf{iLSHADE$_{\epsilon}$} &  6.06E+03 & 6.11E+03 & 8.48E+03 & 1.49E+04 & 3.14E+03 & 100 & 0.00E+00 & 0 \\
                               & \textbf{AHO} & 6.06E+03 & 6.21E+03 & 7.33E+03 & 6.15E+03 & 6.25E+00 & 100 & 0.00E+00 & 22 \\ \hline
\multirow{4}{*}{\textbf{RC19}} & \textbf{IUDE} &  1.67E+00 & 1.67E+00 & 1.67E+00 & 1.67E+00 & 1.20E-16 & 100 & 0.00E+00 & 100 \\
                               & \textbf{$\epsilon$MAgES} &  1.67E+00 & 1.67E+00 & 1.69E+00 & 1.85E+00 & 3.95E-02 & 100 & 0.00E+00 & 44 \\
                               & \textbf{iLSHADE$_{\epsilon}$} &  1.67E+00 & 1.67E+00 & 1.67E+00 & 1.67E+00 & 7.59E-07 & 100 & 0.00E+00 & 100 \\
                               & \textbf{AHO} & 1.67E+00 & 1.67E+00 & 1.67E+00 & 1.67E+00 & 1.15E-14 & 100 & 0.00E+00 & 100 \\ \hline
\multirow{4}{*}{\textbf{RC17}} & \textbf{IUDE} &  1.27E-02 & 1.27E-02 & 1.27E-02 & 1.27E-02 & 1.08E-05 & 100 & 0.00E+00 & 100 \\
                               & \textbf{$\epsilon$MAgES} &  1.27E-02 & 1.27E-02 & 1.27E-02 & 1.37E-02 & 2.16E-04 & 100 & 0.00E+00 & 96 \\
                               & \textbf{iLSHADE$_{\epsilon}$} &  1.27E-02 & 1.27E-02 & 1.30E-02 & 1.78E-02 & 1.06E-03 & 100 & 0.00E+00 & 88 \\
                               & \textbf{AHO} & 1.27E-02 & 1.27E-02 & 1.29E-02 & 1.51E-02 & 2.17E-04 & 100 & 0.00E+00 & 96 \\ \hline
\multirow{4}{*}{\textbf{RC21}} & \textbf{IUDE} &  2.35E-01 & 2.35E-01 & 2.35E-01 & 2.35E-01 & 1.13E-16 & 100 & 0.00E+00 & 100 \\
                               & \textbf{$\epsilon$MAgES} &  2.35E-01 & 2.35E-01 & 2.35E-01 & 2.35E-01 & 1.13E-16 & 100 & 0.00E+00 & 100 \\
                               & \textbf{iLSHADE$_{\epsilon}$} &  2.35E-01 & 2.35E-01 & 2.35E-01 & 2.35E-01 & 1.13E-16 & 100 & 0.00E+00 & 100 \\
                               & \textbf{AHO} & 2.35E-01 & 2.35E-01 & 2.35E-01 & 2.35E-01 & 1.13E-16 & 100 & 0.00E+00 & 100 \\ \hline
\end{tabular}
}
\end{table}

Table \ref{table29} summarizes the performance of AHO on five challenging engineering problems. In addition, the statistical results of AHO are compared to IUDE, $\epsilon$MAgES, and iLSHADE$_{\epsilon}$, which are 3 well-established metaheuristic algorithms \cite{kumar2020test}. The statistical results of metaheuristic algorithms for 25 independent runs are shown in terms of the worst value, the median, the mean, the best value, and the standard deviation value for each optimizer. The following criteria are used to assess the difficulty level of the considered problems.

\begin{itemize}
\item Feasibility Rate (FR): the ratio of the number of runs in which at least one feasible solution is reached within MaxFEs and the total number of runs.
\item Mean constraint Violation (MV): it is calculated using Equation \ref{equation10} \cite{kumar2020test}.
\begin{equation}
\label{equation10}
MV=\frac{\sum_{i=1}^{p}\max(g_i(x),0)+\sum_{i=p+1}^{m}\max(h_i(x)-10^{-4},0)}{m}
\end{equation}
\item Success Rate (SR): the ratio of the total number of runs in which an algorithm has reached a feasible solution $x$ satisfying $f(x)-f(x^*) \leq 10^{-8}$ within MaxFEs and total runs.
\end{itemize}

To compare the performance of IUDE, $\epsilon$MAgES, iLSHADE$_{\epsilon}$, and AHO on the considered constrained engineering design problems, we have used the ranking scheme proposed in \cite{kumar2020test}. Table \ref{table30} gives the ranking of IUDE, $\epsilon$MAgES, iLSHADE$_{\epsilon}$, and AHO based on the performance measure (PM) proposed in \cite{kumar2020test}. We observe that $\epsilon$MAgES and iLSHADE$_{\epsilon}$ have the same rank which means their performance is the same. On the other hand, AHO outperforms only IUDE.

\begin{table}[!]
\centering
\caption{PM values of IUDE, $\epsilon$MAgES, iLSHADE$_{\epsilon}$, and AHO of the considered problems.}
\label{table30}
\begin{tabular}{lll}
\hline
\textbf{Algorithm} & \textbf{PM} & \textbf{Rank} \\ \hline
IUDE & 0.0119 & 4 \\
$\epsilon$MAgES & 0.0116 & 1.5 \\
iLSHADE$_{\epsilon}$ & 0.0116 & 1.5 \\
AHO & 0.0117 & 3 \\ \hline
\end{tabular}
\end{table}

\subsection{Weight Minimization of a Speed Reducer}

It requires the design of a speed reducer for a miniature aircraft engine. The optimization problem has the following form.

Minimize:

\begin{equation*}
\begin{split}
f(x)=0.7854x_2^2x_1(14.9334x_3-43.0934+3.3333x_3^2)+\\0.7854(x_5x_7^2+x_4x_6^2)-1.508x_1(x_7^2+x_6^2)+7.477(x_7^3+x_6^3)
\end{split}
\end{equation*}

Subject to:

\begin{equation*}
\left\{
\begin{array}{l}
g_1(x)=-x_1x_2^2x_3 + 27 \leq 0 \\
g_2(x)=-x_1x_2^2x_3^2 + 397.5 \leq 0 \\
g_3(x)=-x_2x_6^4x_3x_4^{-3} + 1.93 \leq 0 \\
g_4(x)=-x_2x_7^4x_3x_5^{-3} + 1.93 \leq 0 \\
g_5(x)=10x_6^{-3}\sqrt{16.91 \times 10^6 + (745x_4x_2^{-1}x_3^{-1})^2} - 1100 \leq 0 \\
g_6(x)=10x_7^{-3}\sqrt{157.5 \times 10^6 + (745x_5x_2^{-1}x_3^{-1})^2} - 850 \leq 0 \\
g_7(x)=x_2x_3 - 40 \leq 0 \\
g_8(x)=-x_1x_2^{-1} + 5 \leq 0 \\
g_9(x)=x_1x_2^{-1} - 12 \leq 0 \\
g_{10}(x)=1.5x_6 - x_4 + 1.9 \leq 0 \\
g_{11}(x)=1.1x_7 - x_5 + 1.9 \leq 0 \\
\end{array}
\right.
\end{equation*}

With bounds:

\begin{equation*}
\left\{
\begin{array}{l}
2.6 \leq x_1 \leq 3.6 \\
0.7 \leq x_2 \leq 0.8 \\
17 \leq x_3 \leq 28 \\
7.3 \leq x_4 \leq 8.3 \\
7.3 \leq x_5 \leq 8.3 \\
2.9 \leq x_6 \leq 3.9 \\
5 \leq x_7 \leq 5.5 \\
\end{array}
\right.
\end{equation*}

\subsection{Pressure vessel design}

The objective of this problem is to optimize the welding cost, material, and forming of a vessel. This problem has four constraints to be satisfied, and four variables are used to compute the objective function value: shell thickness ($z_1$), head thickness ($z_2$), inner radius ($x_3$), and length of the vessel without including the head ($x_4$). This problem can be formulated as follows.

Minimize:

\begin{equation*}
f(x)=1.7781z_2x_3^2 + 0.6224z_1x_3x_4 + 3.1661z_1^2x_4 + 19.84z_1^2x_3
\end{equation*}

Subject to:

\begin{equation*}
\left\{
\begin{array}{l}
g_1(x)=0.00954x_3 \leq z_2 \\
g_2(x)=0.0193x_3 \leq z_1 \\
g_3(x)=x_4 \leq 240 \\
g_4(x)=-\pi x_3^2x_4 - \frac{4}{3}\pi x_3^3 \leq -1296000 \\
\end{array}
\right.
\end{equation*}

Where:

\begin{equation*}
\left\{
\begin{array}{l}
z_1=0.0625x_1 \\
z_2=0.0625x_2 \\
\end{array}
\right.
\end{equation*}

With bounds:

\begin{equation*}
\left\{
\begin{array}{l}
x_1 \in \{1,\ldots,99\} \\
x_2 \in \{1,\ldots,99\} \\
10 \leq x_3 \leq 200 \\
10 \leq x_4 \leq 200 \\
\end{array}
\right.
\end{equation*}

\subsection{Welded beam design}

The objective of this problem is to design a welded beam with minimum cost. This problem has five constraints, and four variables. The mathematical description of this problem can be defined as follows.

Minimize:

\begin{equation*}
f(x)=0.04811x_3x_4(x_2 + 14) + 1.10471x_1^2x_2
\end{equation*}

Subject to:

\begin{equation*}
\left\{
\begin{array}{l}
g_1(x)=x_1 - x_4 \leq 0 \\
g_2(x)=\delta(x) - \delta_{\max} \leq 0 \\
g_3(x)=P \leq P_c(x) \\
g_4(x)=\tau_{\max}\geq \tau(x) \\
g_5(x)=\sigma(x) - \sigma_{\max} \leq 0 \\
\end{array}
\right.
\end{equation*}

Where

\begin{equation*}
\left\{
\begin{array}{l}
\tau=\sqrt{\tau'^2 + \tau''^2 + 2\tau'\tau''^2\frac{x_2}{2R}} \\
\tau''=\frac{RM}{J} \\
\tau'=\frac{P}{\sqrt{2}x_2x_1} \\
M=p(\frac{x_2}{2} + L) \\
R=\sqrt{\frac{x_2^2}{4} + (\frac{x_1 +x_3}{2})^2} \\
J=2 ((\frac{x_2^2}{4} + (\frac{x_1 + x_3}{2})^2)\sqrt{2}x_1x_2) \\
\sigma(x)=\frac{6PL}{x_4x_3^2} \\
\delta(x)=\frac{6PL^3}{Ex_3^2x_4}  \\
P_c(x)=\frac{4.013Ex_3x_4^3}{6L^2} (1 - \frac{x_3}{2L}\sqrt{\frac{E}{4G}}) \\
L=14in \\
P=6000lb \\
E=30 \times 10^6psi \\
\sigma_{\max}=30, 000psi \\
\tau_{\max}=13, 600psi \\
G=12.106psi \\
\delta_{\max}=0.25in \\
\end{array}
\right.
\end{equation*}

With bounds:

\begin{equation*}
\left\{
\begin{array}{l}
0.125 \leq x_1 \leq 2 \\
0.1 \leq x_2 \leq 10 \\
0.1 \leq x_3 \leq 10 \\
0.1 \leq x_4 \leq 2 \\
\end{array}
\right.
\end{equation*}

\subsection{Tension/compression spring design}

The objective of this problem is to optimize the weight of a tension or compression spring. This problem contains four constraints and three variables: the diameter of the wire ($x_1$), the mean of the diameter of coil ($x_2$), and the number of active coils ($x_3$). This problem is defined in the following way.

Minimize:

\begin{equation*}
f(x)=x_1^2x_2(2 + x_3)
\end{equation*}

Subject to:

\begin{equation*}
\left\{
\begin{array}{l}
g_1(x)=1 - \frac{x_2^3x_3}{71785x_1^4}\leq 0 \\
g_2(x)=\frac{4x_2^2 - x_1x_2}{12566(x_2x_1^3 - x_1^4)}+\frac{1}{5108 x_1^2} + - 1 \leq 0 \\
g_3(x)=1 - \frac{140.45x_1}{x_2^2x_3}\leq 0 \\
g_4(x)=\frac{x_1 + x_2}{1.5}- 1 \leq 0 \\
\end{array}
\right.
\end{equation*}

With bounds:

\begin{equation*}
\left\{
\begin{array}{l}
0.05 \leq x_1 \leq 2.00 \\
0.25 \leq x_2 \leq 1.30 \\
2.00 \leq x_3 \leq 15.0 \\
\end{array}
\right.
\end{equation*}

\subsection{Multiple disk clutch brake design problem}

The objective of this problem is to minimize the mass of a multiple disk clutch brake. This problem has five integer decision variables: the inner radius ($x_1$), the outer radius ($x_2$), the disk thickness ($x_3$), the force of actuators ($x_4$), and the number of frictional surfaces ($x_5$). This problem contains 8 non-linear constraints. The problem can be defined as follows.

Minimize:

\begin{equation*}
f(x)=\pi(x_2^2 - x_1^2)x_3(x_5 + 1)
\end{equation*}

Subject to:

\begin{equation*}
\left\{
\begin{array}{l}
g_1(x)=-p_{\max} + p_{rz} \leq 0 \\
g_2(x)=p_{rz}V_{sr} - V_{sr,\max}p_{\max} \leq 0 \\
g_3(x)=\Delta R + x_1 - x_2 \leq 0 \\
g_4(x)=-L_{\max} + (x_5 + 1)(x_3 + \delta) \leq 0 \\
g_5(x)=sM_{s} - M_{h} \leq 0 \\
g_6(x)=T \geq 0 \\
g_7(x)=-V_{sr,\max}p_{\max} + V_{sr} \leq 0 \\
g_8(x)=T - T_{\max} \leq 0 \\
\end{array}
\right.
\end{equation*}

Where

\begin{equation*}
\left\{
\begin{array}{l}
M_{h}=\frac{2}{3}\mu x_4x_5\frac{x_2^3 - x_1^3}{x_2^2 - x_1^2} N.mm \\
\omega=\frac{\pi n}{30}rad/s \\
A=\pi(x_2^2 - x_1^2)mm^2 \\
p_{rz}=\frac{x_4}{A}N/mm2 \\
V_{sr}=\frac{\pi R_{sr}n}{30}mm/s \\
R_{sr}=\frac{2}{3}\frac{x_2^3 - x_1^3}{x_2^2x_1^2} mm \\
T=\frac{I_{z}\omega}{M_{h} + M_{f}} \\
\Delta R=20mm \\
L_{\max}=30mm \\
\mu=0.6 \\
V_{sr,\max}=10m/s\\
\delta=0.5mm \\
s=1.5 \\
T_{\max}=15s \\
n=250rpm \\
I_{z}=55Kg.m^2 \\
M_{s}= 40Nm \\
M_{f}=3Nm \\
p_{\max}=1 \\
\end{array}
\right.
\end{equation*}

With bounds:

\begin{equation*}
\left\{
\begin{array}{l}
x_1 \in \{60,\ldots,80\} \\
x_2 \in \{90,\ldots,110\} \\
x_3 \in \{1,\ldots,3\} \\
x_4 \in \{0,\ldots,1000\} \\
x_5 \in \{2,\ldots,9\} \\
\end{array}
\right.
\end{equation*}

\section{Conclusion and future work}
\label{section6}

In this paper, a new population-based optimization algorithm termed the Archerfish Hunting Optimizer is introduced to handle constrained and unconstrained optimization problems. AHO is founded on the shooting and jumping behaviors of archerfish when hunting aerial insects in nature. Some equations are outlined to model the hunting behavior of archerfish to solve optimization problems. Ten unconstrained optimization problems are used to assess the performance of AHO. The exploration, exploitation, and local optima avoidance's capacities are examined using unimodal, basic, hybrid, and composition functions. The obtained statistical results of the Wilcoxon signed-rank and the Friedman tests confirm that AHO can find superior solutions by comparison to 12 well-acknowledged optimizers. AHO's collected numerical outcomes on three constrained engineering design problems also show that AHO has outstanding results compared to 3 well-regarded optimizers.

AHO is straightforwardly explained with uncomplicated exploration and exploitation techniques. It is desirable to introduce other evolutionary schemes such as mutation, crossover, or multi-swarm, which we plan to do in the future. In addition, we plan to develop the binary and multi-objective versions of AHO.

\section*{Acknowledgement}

This research work is supported by UAEU Grant: 31T102-UPAR-1-2017. We appreciate the constructive comments of anonymous reviewers.

\nocite{*}
\bibliographystyle{fundam}
\bibliography{citations}


\end{document}